\documentclass[11pt]{article}

\usepackage[preprint]{acl}

\usepackage{times}
\usepackage{latexsym}
\usepackage{amsmath}

\usepackage[T1]{fontenc}

\usepackage[utf8]{inputenc}

\usepackage{microtype}

\usepackage{inconsolata}

\usepackage{graphicx}

\usepackage{booktabs}
\usepackage{multirow}
\usepackage{array}
\usepackage{xspace} 
\usepackage[nolist,nohyperlinks]{acronym}

\usepackage{lipsum}  
\usepackage{amsfonts}

\usepackage{geometry} 
\usepackage[rightcaption]{sidecap} 

\usepackage{tikz}
\usetikzlibrary{shapes.geometric, arrows.meta, positioning}
\usepackage{amsmath}
\usepackage{algorithm}
\usepackage{algorithmic}
\usepackage{listings}
\usepackage{enumitem} 
\usepackage{lipsum}
\usepackage[most,listings,breakable]{tcolorbox}
\definecolor{thinkcolor}{RGB}{227,196,144}
\definecolor{observecolor}{RGB}{153,201,227}
\definecolor{explorecolor}{RGB}{178,217,200}
\definecolor{prompt1_bg}{RGB}{252, 248, 245}  
\definecolor{prompt1_frame}{RGB}{191, 97, 106}  
\definecolor{original_box}{RGB}{255, 250, 240} 
\definecolor{compressed_box}{RGB}{240, 248, 255} 
\definecolor{action_color}{RGB}{0, 0, 0}  
\definecolor{purple_color}{RGB}{163, 73, 164}  

\newcounter{promptexample}[section]
\renewcommand{\thepromptexample}{\arabic{promptexample}}

\usepackage{array, multirow, tabularx, booktabs, makecell}
\usepackage{svg}

\title{\approach: Vision-Free Navigation Instruction Evaluation via Graph Reasoning on OpenStreetMap}

\author{
 \textbf{Farzad Shami\textsuperscript{1}},
 \textbf{Subhrasankha Dey\textsuperscript{1}},
 \textbf{Nico Van de Weghe\textsuperscript{2}},
 \textbf{Henrikki Tenkanen\textsuperscript{1}}
\\
\\
 \textsuperscript{1}Aalto University,
 \textsuperscript{2}Ghent University
\\
\small{
  \texttt{\{\href{mailto:farzad.shami@aalto.fi}{farzad.shami},\href{mailto:subhrasankha.dey@aalto.fi}{subhrasankha.dey},\href{mailto:henrikki.tenkanen@aalto.fi}{henrikki.tenkanen}\}@aalto.fi, \href{mailto:nico.vandeweghe@ugent.be}{nico.vandeweghe}@ugent.be}
  }
}

\begin{document}
\begin{acronym}
    \acro{NLP}{Natural Language Processing}
    \acro{LLM}{Large Language Model}
    \acro{OSM}{Open Street Map}
    \acro{POI}{Point of Interest}
    \acro{NER}{Named Entity Recognition}
    \acro{CoT}{Chain of Thought}
    \acro{BFS}{Breadth-First Search}
    \acro{VLN}{Vision-and-Language Navigation}
    \acro{HRI}{Human-Robot Interaction}
    \acro{NLG}{Natural Language Generation}
    \acro{VLA}{Vision-Language-Action}
\end{acronym}

\maketitle
\begin{abstract}
The evaluation of navigation instructions remains a persistent challenge in Vision-and-Language Navigation (VLN) research. Traditional reference-based metrics such as BLEU and ROUGE fail to capture the functional utility of spatial directives, specifically whether an instruction successfully guides a navigator to the intended destination. 
Although existing VLN agents could serve as evaluators, their reliance on high-fidelity visual simulators introduces licensing constraints and computational costs, and perception errors further confound linguistic quality assessment.
This paper introduces \approach (\textbf{G}raph-based \textbf{R}easoning over \textbf{O}SM \textbf{K}nowledge for instruction \textbf{E}valuation), a vision-free training-free hierarchical LLM-based framework for evaluating navigation instructions using OpenStreetMap data. Through systematic ablation studies, we demonstrate that structured JSON and textual formats for spatial information substantially outperform grid-based and visual graph representations. Our hierarchical architecture combines sub-instruction planning with topological graph navigation, reducing navigation error by 68.5\% compared to heuristic and sampling baselines on the Map2Seq dataset. The agent's execution success, trajectory fidelity, and decision patterns serve as proxy metrics for functional navigability given OSM-visible landmarks and topology, establishing a scalable and interpretable evaluation paradigm without visual dependencies. Code and data are available at \url{https://anonymous.4open.science/r/groke}.
\end{abstract}

\section{Introduction}
\paragraph{Evaluation Deficits in Embodied Systems.} The intersection of \ac{NLP} and robotics has given rise to the field of Embodied AI~\cite{tellex2020robots, duan2022survey, anderson2018vision}, where agents must perceive, reason, and act within physical environments based on linguistic directives~\cite{cong2025overview, gu2022vision}. Within this domain, \ac{VLN} has emerged as a flagship task, requiring agents to navigate complex 3D environments following natural language instructions~\cite{krantz2020beyond,ku2020room}. While significant progress has been made in developing agents that can follow instructions -- evidenced by performance gains on benchmarks like Room-to-Room (R2R)~\cite{Anderson_2018_CVPR} and Touchdown~\cite{chen2019touchdown} -- a critical and often overlooked component of this ecosystem is the evaluation of the instructions themselves. This paper addresses the evaluation of navigation instructions rather than the navigation agents that execute them, focusing on whether instructions themselves are clear and navigable.

\begin{figure}
    \centering
    \includegraphics[width=\linewidth]{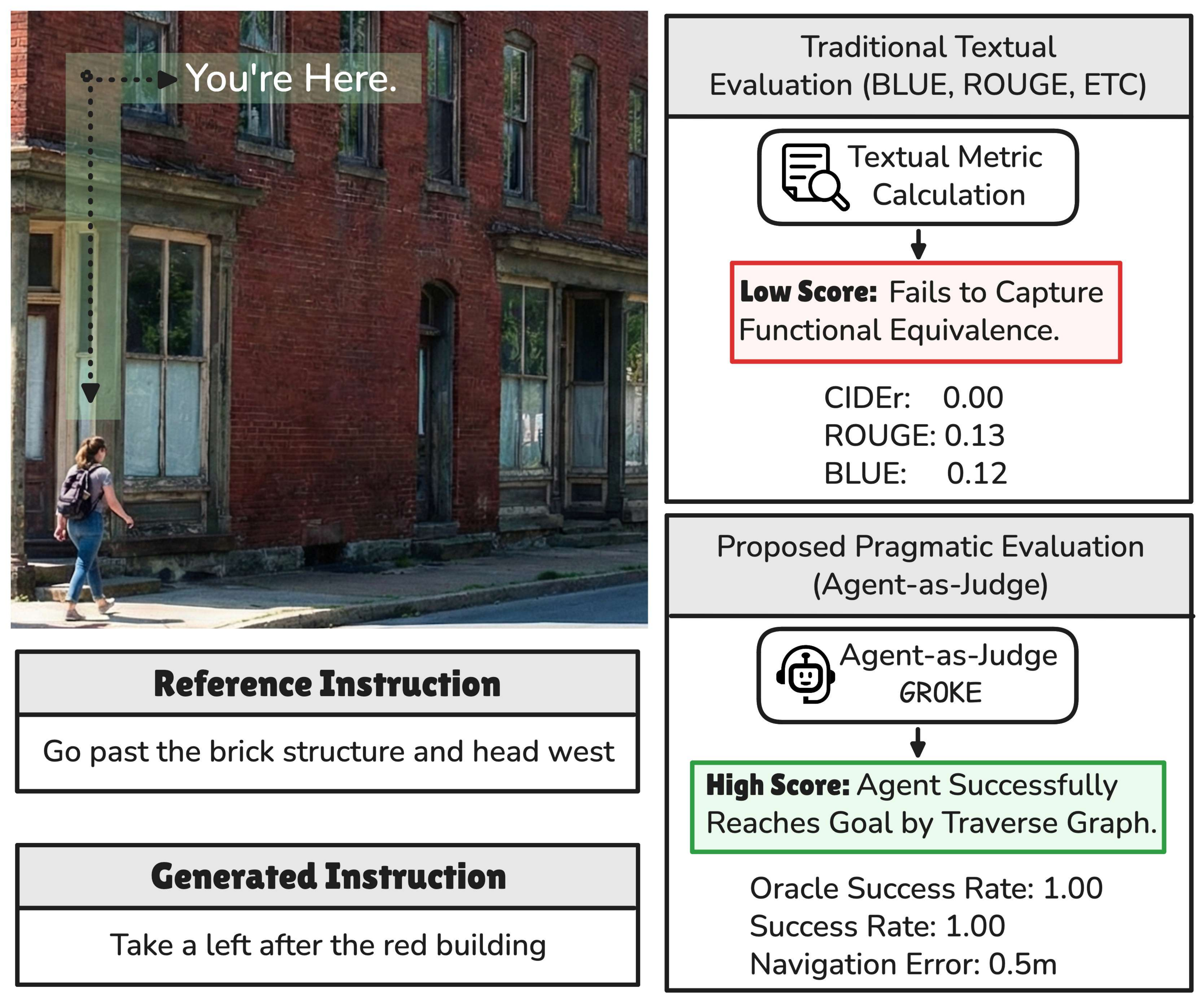}
    \caption{Comparison of traditional textual metrics and proposed pragmatic evaluation metrics.}
    \label{fig:comparion}
\end{figure}

Historically, the navigability of navigation instructions, whether generated by humans or automated ``Speaker'' models, has been assessed using reference-based metrics adapted from machine translation and image captioning. Metrics such as BLEU~\cite{papineni2002bleu}, ROUGE~\cite{lin2004rouge}, METEOR~\cite{banerjee2005meteor}, and CIDEr~\cite{vedantam2015cider} calculate n-gram overlaps between a candidate instruction and a ``gold standard'' reference instruction. This methodology rests on the precarious assumption that there is a single correct way to describe a route and that lexical similarity correlates with functional utility. 

However, in the context of spatial navigation, this assumption breaks down~\cite{zhao2021evaluation, jain2019stay}. An instruction that states \textit{``Turn left at the bank''} shares high lexical overlap with \textit{``Turn right at the bank,''} yet the functional outcome of following these two directives is diametrically opposed. Conversely, \textit{``Take a left after the red building''} and \textit{``Go past the brick structure and head west''} may share zero n-grams but describe the exact same valid action. 

The ``meaning'' of a navigation instruction is not defined by its syntax but by its compliance conditions -- the set of physical trajectories that satisfy the directive~\cite{ilharco2019general}. This disconnect creates a bottleneck in the advancement of \ac{HRI}~\cite{fong2004common}. If we cannot accurately measure the navigability of an instruction, we cannot train systems to generate better ones, nor can we filter low-quality human data from training sets. Previous work~\cite{chenalpagasus} shows that automatically filtering low-quality data can improve model training in general domains. Addressing this requires a shift from intrinsic, text-based evaluation to extrinsic, pragmatic evaluation, where the navigability of an instruction is measured by the success of a rational agent attempting to follow it.

\paragraph{Visual Reliance in Pragmatic Evaluation.} Existing attempts to implement pragmatic evaluation typically involve training a ``follower'' agent~\cite{fried2018speaker, anderson2018evaluation} to execute instructions in a simulator and measuring its success rate. If the agent succeeds, the instruction is deemed good; if it fails, the instruction is deemed bad. While theoretically sound, the practical implementation of this approach has been heavily reliant on high-fidelity visual simulators like Matterport3D or Google Street View (GSV)~\cite{anderson2018vision, chang2017matterport3d, schumann-riezler-2021-map2seq}.

This reliance on photorealistic visual inputs introduces several critical issues. First, it conflates linguistic quality with visual recognition capabilities~\cite{wang2019reinforced}. If an evaluation agent fails to execute an instruction because it cannot identify a ``stucco wall'' in a grainy panoramic image, the failure is attributed to the instruction, even if the text was perfectly clear. This introduces noise into the evaluation metric, making it difficult to isolate the quality of the \ac{NLG} from the agent's computer vision performance. Second, the reliance on proprietary datasets like GSV creates significant barriers to reproducibility~\footnote{\url{https://about.google/intl/en-GB_ALL/brand-resource-center/products-and-services/geo-guidelines/}} and scalability for out-door navigation. Researchers must grapple with API costs, licensing restrictions, and the sheer data volume of downloading terabytes of panoramic imagery. This effectively limits the accessibility of pragmatic evaluation to well-funded labs and hinders the democratization of \ac{VLN} research.

In addition, previous work shows that \ac{VLA} models fail in clutter not because control is weak, but because perception collapses~\cite{vo2025clutter}. In real scenes, they over-grasp, chase distractors, and act even when the target is absent.

The Map2Seq dataset~\cite{schumann-riezler-2021-map2seq} presents a unique opportunity to address these limitations. Unlike Touchdown, which is inextricably linked to GSV panoramas, map2seq provides aligned \ac{OSM} data -- nodes, edges, and \acp{POI} -- alongside navigation routes. This allows for a decoupling of the visual and linguistic modalities. By abstracting the environment into a semantic representation derived from \ac{OSM}, we can build evaluation agents that reason over map features rather than raw pixels. This not only removes the visual recognition bottleneck but also allows for the evaluation of instructions based on their structural and semantic logic, which is arguably the most critical aspect of human-generated navigation guidance.

\paragraph{Research Objectives.} 
This research presents a novel methodological approach to evaluate navigation instructions.
Rather than asking ``How well did the agent perform?'' we ask ``How navigable is this instruction?'' This inversion of the standard \ac{VLN} problem formulation allows us to treat the agent's performance metrics (execution success, trajectory fidelity, and decision entropy) as proxy scores for the navigability of the input text. Our specific research objectives are as follows:

First, we aim to develop a visually agnostic environmental representation by transforming the sparse vector data of Map2Seq into structured spatial representations that capture topological connectivity, geometric layout, and semantic landmarks. 
Through systematic comparison of multiple encoding strategies (i.e., textual narratives, structured JSON graphs, grid-based matrices, and GraphViz-style visualizations), we identify which representation formats enable optimal LLM reasoning for navigation tasks. Our ablation studies, investigate whether dense grid rasterization or structured textual schemas provide better foundation for spatial reasoning in language models.

Second, we aim to design a hierarchical agent inspired by embodied AI and construct an agent that separates high-level instruction parsing from low-level path planning. The Sub-instruction Agent decomposes natural language instructions into atomic sub-goals and extracts landmark references, while the Navigator Agent executes these sub-goals through iterative waypoint selection. This hierarchical decomposition reduces the complexity of long-horizon navigation by creating manageable action sequences that can be tracked and validated independently.

Third, we establish a suite of evaluation metrics for instruction navigability, we adapt standard \ac{VLN} metrics (Navigation Error, Success Rate, Oracle Success Rate, and Success weighted by Dynamic Time Warping) to serve as inverse indicators of instruction navigability. When an agent successfully follows an instruction with low navigation error and high path fidelity, this suggests the instruction itself is well-formed and navigable. Conversely, high error rates and trajectory deviations may indicate ambiguous or underspecified instructions, even when the ground truth path is known.

\paragraph{Contribution.}
We propose \approach (\textbf{G}raph-based \textbf{R}easoning over \textbf{O}SM \textbf{K}nowledge for instruction \textbf{E}valuation), a modular system for the agentic evaluation of navigation instructions. The contributions are three-fold.

\begin{itemize}
\item[(1)] \approach systematically compares four spatial representation formats, and demonstrate that structured JSON significantly outperform  representations in this task. Our ablation studies reveal that JSON format achieves Navigation Error of 41.3 (m) and Success Rate of 74.0\%.

\item[(2)] \approach formalizes the Agent-as-Judge methodology, providing an experimental protocol to validate the evaluator itself. We address the ``meta-evaluation'' problem -- how to ensure the judge is accurate -- by correlating agentic metrics with human judgments of instruction clarity and by performing studies on map representations and instruction segmentation.

\item[(3)] Within \approach, we provide a detailed implementation of this framework, including specific algorithmic descriptions for graph serialization, prompt engineering strategies for hierarchical reasoning, and statistical methods for analyzing agent trajectories. 
\end{itemize}

The rest of the paper is organized as follows. In Section~\ref{sec:related_work}, we review related work in navigation instruction evaluation and \ac{VLN}. In Section~\ref{sec:method}, we introduce the \approach agent architecture and its core components. We detail the \approach construction in Section~\ref{sec:experiment}, covering dataset preparation and implementation details alongside experimental results. Finally, in Section~\ref{sec:final_remarks}, we draw some final remarks. 
Section~\ref{sec:appendix} presents systematic ablation studies examining spatial representation formats, sub-instruction planning effectiveness, \ac{POI} detection accuracy, and the impact of \ac{LLM} thinking procedures on navigation performance.

\section{Background \& Related Work}\label{sec:related_work}
\paragraph{Transition from Vision-Dependent to Graph-Based Navigation.}
The task of \ac{VLN} was established by the Room-to-Room (R2R)~\cite{Anderson_2018_CVPR} benchmark, which defined the task of grounding natural language instructions into actions within widely used Matterport3D environments. While subsequent datasets like Touchdown~\cite{chen2019touchdown, mehta2020retouchdown} extended this formulation to outdoor urban settings using Google Street View panoramas, these vision-dependent approaches face significant scalability challenges. Specifically, reliance on high-resolution imagery introduces critical barriers related to privacy regulations, incomplete geographic coverage, and the temporal outdatedness of static captures. Consequently, recent research has pivoted toward map-based methodologies~\cite{schumann2022analyzing}. Work such as Map2Seq~\cite{schumann-riezler-2021-map2seq} has demonstrated that structured geographic data from \ac{OSM} can effectively substitute for visual imagery. By encoding routes into location-invariant graph representations, these approaches enable scalable navigation instruction tasks without the computational overhead or privacy constraints inherent to visual processing.

\paragraph{Spatial Representation for Language Model Reasoning}
The effectiveness of \acp{LLM} in navigation depends on spatial encoding format. Studies indicate structured textual representations often rival visual inputs for spatial reasoning. For example, STMR~\cite{gao2024aerial} substantially reduced Navigation Error by combining semantic labels with topological connectivity. Furthermore, ``Talk like a Graph''~\cite{fatemi2024talk} reveals that encoding choices, such as incident encoding, can boost performance significantly for graph reasoning problems. This is supported by findings from MapGPT~\cite{chen2024mapgpt}, where online-constructed linguistic topological maps enabled ``global-view'' reasoning that purely local visual observations could not support. These results validate the use of symbolic formats, such as JSON-encoded \ac{OSM} data, as a robust alternative to visual perception for driving \ac{LLM}-based agents~\cite{xie2024empowering}.

\paragraph{Agent-Based Evaluation and Metric Evolution}
The evaluation of navigation instructions has historically relied on reference-based text metrics like BLEU; however, even with available references, these metrics often fail to capture the functional utility required for successful navigation~\cite{zhao2021evaluation}. To address this, the field has adopted an agent-based evaluation philosophy, exemplified by the Speaker-Follower paradigm, where instruction quality is measured by the navigational success of a follower agent. This shift requires metrics that capture both goal completion and trajectory fidelity. Although success weighted by Path Length (SPL) penalizes inefficient exploration, it allows agents to take shortcuts that ignore instructions~\cite{jain2019stay}. Therefore, metrics such as normalized Dynamic Time Warping (nDTW) and Success weighted by DTW (SDTW) are essential for ensuring that agents adhere to the intended path. Contemporary hierarchical architectures, such as those seen in VELMA~\cite{schumann2024velma} and NavGPT~\cite{zhou2024navgpt}, leverage these insights by separating high-level reasoning from low-level execution, confirming that functional navigation performance is the most reliable proxy for instruction quality.

\section{Methodology}\label{sec:method}
\begin{figure*}[t!]
    \centering
    \includegraphics[height=5.1cm]{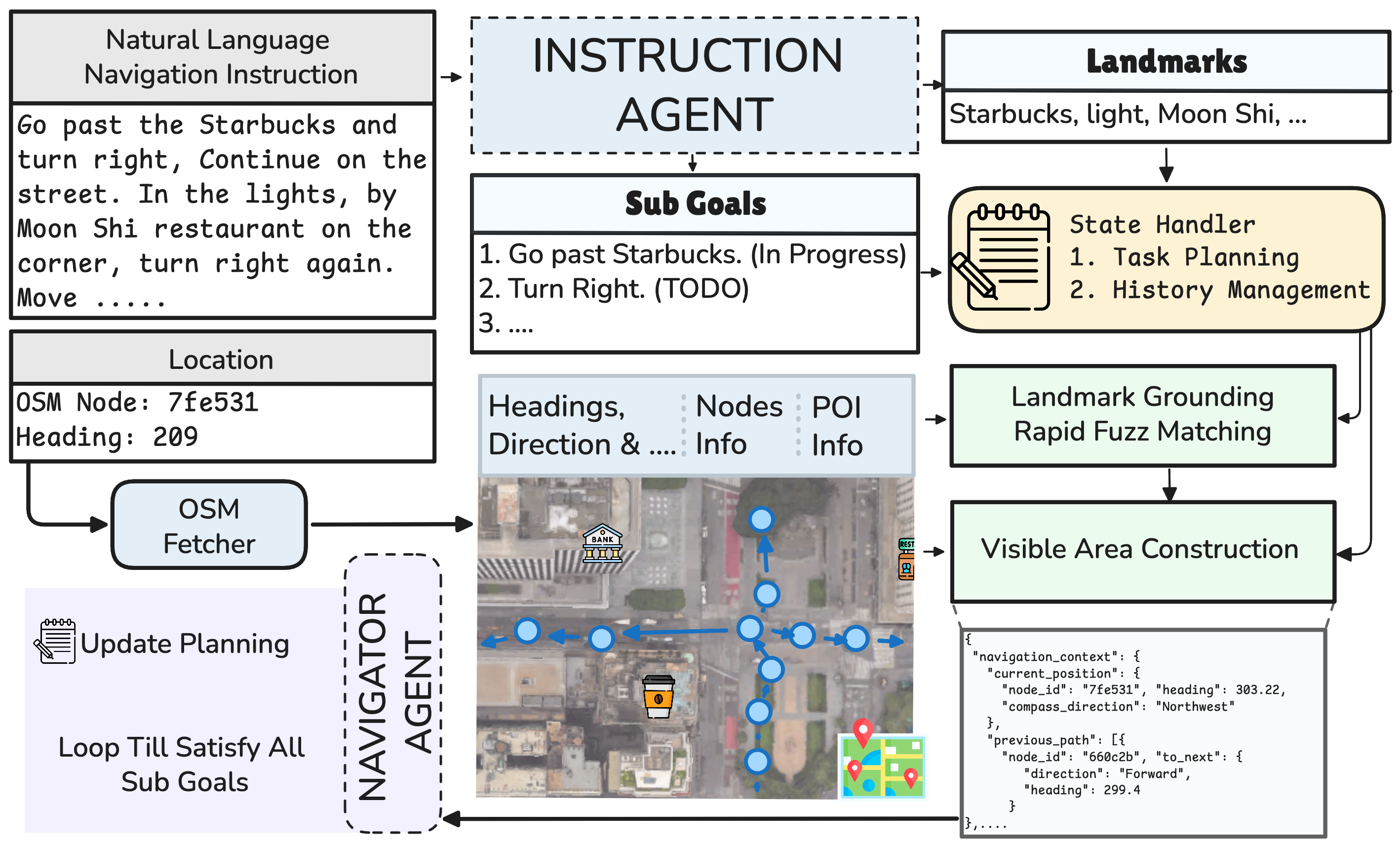}%
    \hspace{0.1em}
    \vrule width 0.4pt
    \hspace{0.1em}
    \includegraphics[height=5.1cm]{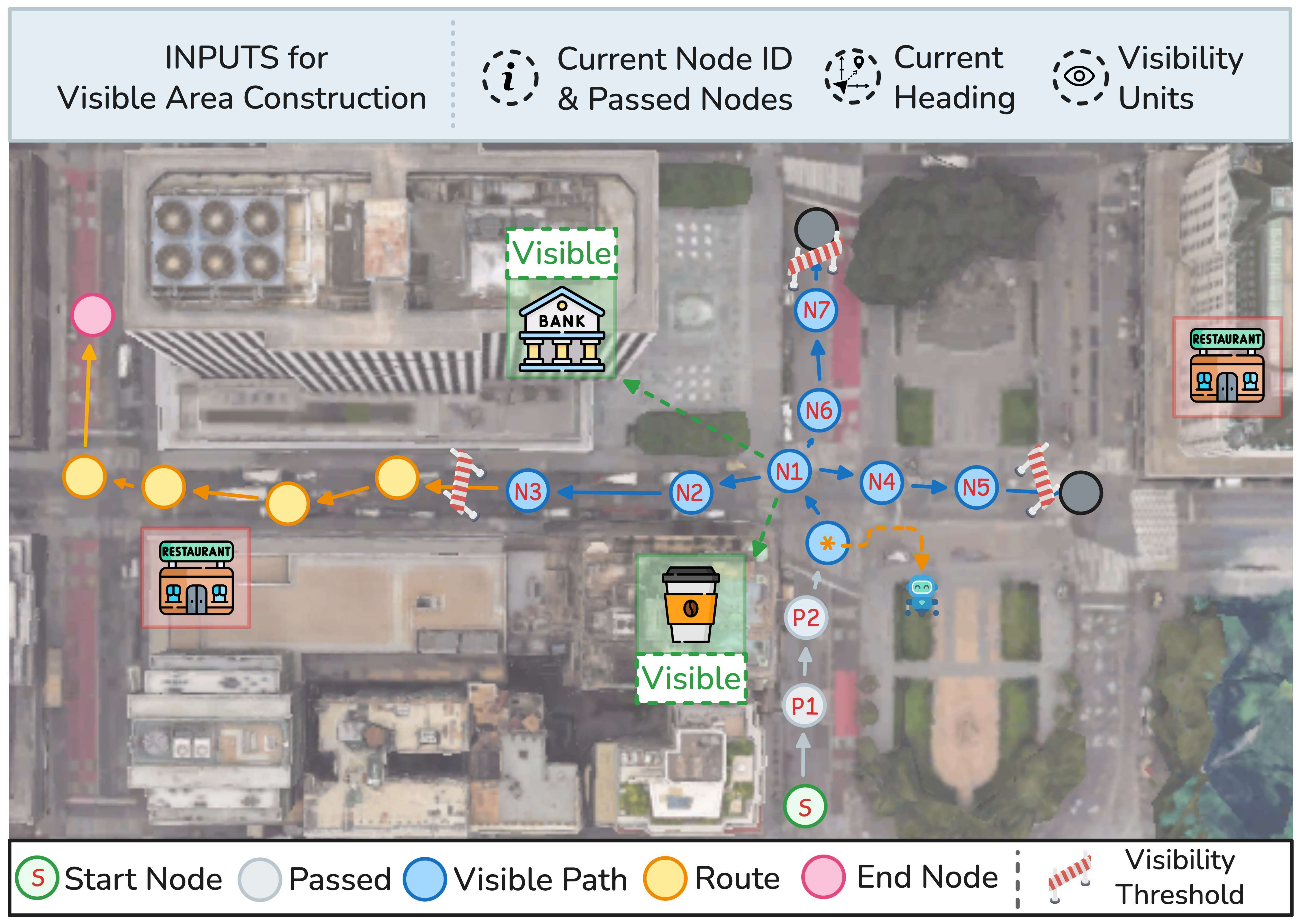}%
    \caption{(Left) \approach consists of three modules: Sub-Goal \& \ac{POI} Extraction, Visible Area Construction, and Navigator Agent. These modules generate sub-goal instructions, construct spatial information representations for the surrounding area, and traverse the graph, respectively. (Right) Visible Area Construction in detail: The map highlights the distinction between different kinds of nodes for immediate local context construction. It demonstrates how the system filters out distant data using visibility thresholds to construct the immediate navigable context.}
    \label{fig:overall_architecture}
\end{figure*}

In this section, we explore how navigation performance can be enhanced by structuring the decision space through hierarchical instruction decomposition and structured spatial representations. We begin with an overview of the navigation task and introduce our proposed agentic system for action prediction ($\S$~\ref{sec:overview}). We then present two core modules of our approach: (1) the Sub-instruction Agent for instruction parsing and landmark extraction ($\S$~\ref{sec:subinstruction}), and (2) the Navigator Agent for spatial reasoning and waypoint selection ($\S$~\ref{sec:navigator}).

\subsection{Overview}
\label{sec:overview}

\paragraph{Problem Formulation.} 
We formalize the task of \ac{VLN} on \ac{OSM} as a sequential decision-making problem in a graph-structured environment. Let $I = \{w_1, w_2, \ldots, w_L\}$ be a natural language instruction consisting of $L$ words. Let $\mathcal{G} = (V, E, P)$ represent the \ac{OSM} graph, where $V$ denotes the set of navigable nodes (street intersections and waypoints), $E$ denotes the set of directed edges connecting nodes with associated headings, and $P$ denotes the set of \acp{POI} with spatial coordinates and semantic tags. Let $\tau^* = \{v_1, v_2, \ldots, v_T\}$ be the ground-truth navigation trajectory, represented as a sequence of nodes $v_t \in V$ from the starting location $v_1$ to the goal location $v_T$.

The navigation task is to find a policy $\pi$ that predicts the next waypoint $v_{t+1}$ given the instruction $I$, the current position $v_t$, the current heading $h_t \in [0, 360)$ degrees, and the local map context $\mathcal{G}_t$ visible from the current position:
$$\pi: (I, v_t, h_t, \mathcal{G}_t) \rightarrow v_{t+1}$$

The agent's objective is to maximize the probability of reaching the goal location while following the instruction accurately:
$$\pi^* = \arg\max_{\pi} P(\tau_{agent} \approx \tau^* | I, \mathcal{G})$$

where $\tau_{agent} = \{v_1, \pi(I, v_1, h_1, \mathcal{G}_1), \ldots\}$ is the predicted trajectory generated by the agent's policy.

\paragraph{System Architecture.} 
We implement a training-free multi-agent hierarchical system (i.e., \approach) that decomposes the navigation problem into two stages (Figure~\ref{fig:overall_architecture}, left). First, a Sub-instruction Agent parses the full instruction $I$ into a sequence of sub-goals $\{g_1, g_2, \ldots, g_K\}$ and extracts landmark references. Second, a Navigator Agent iteratively executes each sub-goal by reasoning over structured spatial representations of the local environment. The Navigator Agent continues executing the current sub-goal until completion, then advances to the next sub-goal. This hierarchical decomposition reduces the complexity of long-horizon navigation by creating manageable action sequences.

At each navigation step, the system constructs a local spatial context $\mathcal{G}_t$ by extracting the visible area from the current position. This visible area includes all nodes, edges, and \acp{POI} within a defined visibility range (measured in intersection units). The Navigator Agent receives this spatial information in structured formats and predicts the next waypoint $v_{t+1}$ that advances progress toward completing the current sub-goal.

\subsection{Sub-instruction Agent}
\label{sec:subinstruction}

The Sub-instruction Agent transforms the natural language instruction $I$ into structured, machine-readable components that can be executed sequentially. This agent performs two primary functions: (1) decomposing the instruction into atomic sub-goals, and (2) extracting and categorizing landmark references.

\paragraph{Sub-goal Decomposition.}
We define a sub-goal $g_k$ as an atomic navigation action that corresponds to one of three primitive operations: \texttt{MOVE\_FORWARD} (proceed along the current path), \texttt{TURN\_LEFT} (change heading by approximately 90$^\circ$ counterclockwise), or \texttt{TURN\_RIGHT} (change heading by approximately 90$^\circ$ clockwise). The agent uses an \ac{LLM}-based parser to segment the instruction $I$ into a sequence of sub-goals:
$$I \xrightarrow{\text{parse}} \{g_1, g_2, \ldots, g_K\}$$

Each sub-goal $g_k$ is associated with a natural language description (e.g., \textit{``Go straight to the bank''}) and a status indicator (\texttt{IN\_PROGRESS}, \texttt{COMPLETED}, or \texttt{TODO}). The decomposition follows spatial and temporal relationships expressed in the instruction, preserving the sequential order of the actions.

\paragraph{Landmark Extraction and Grounding.}
The Sub-instruction Agent identifies all landmark references in $I$ and categorizes them according to their semantic type. We define the landmarks $\mathcal{L} = \{l_1, l_2, \ldots, l_M\}$ as physical entities mentioned in the instruction, where each landmark $l_m$ has a name and a category (amenity, traffic control, etc.).
After extraction, we perform landmark grounding to associate each instruction landmark with its corresponding \ac{POI} nodes on the map graph $\mathcal{G}$. We use fuzzy string matching with the RapidFuzz library~\cite{max_bachmann_2025_15133267} to match landmark names to \ac{POI} tags:
$$\text{sim}(l_m, p_i) = \text{partial\_ratio}(\text{name}(l_m), \text{tags}(p_i))$$

where $p_i \in P$ is a \ac{POI} on the map. A match is established when $\text{similarity}(l_m, p_i) > \tau$, where $\tau$ represents the similarity threshold. Each grounded landmark is assigned a unique letter identifier (A, B, C, etc.) for reference in the navigation prompts.

\subsection{Navigator Agent}
\label{sec:navigator}

The Navigator Agent executes the sequence of sub-goals by iteratively selecting waypoints that advance progress toward sub-goal completion. At each step $t$, the agent receives the current sub-goal $g_k$, the current position $v_t$, the current heading $h_t$, and the visible area $\mathcal{G}_t$. The agent outputs a prediction:
$$\text{output}_t = (\text{status}_k, v_{t+1})$$

where $\text{status}_k \in \{\texttt{IN\_PROGRESS}, \texttt{COMPLETED}\}$ indicates whether sub-goal $g_k$ has been satisfied, and $v_{t+1} \in V$ is the next waypoint to navigate toward.

\paragraph{Visible Area Construction.}
The visible area $\mathcal{G}_t$ represents the spatial context perceived from the current position. In our implementation, this construction simulates human navigation by defining visibility as the forward view along the street to the next intersection. We model this visible area by traversing the current heading direction to the next topological node in the street network.

Given the current node $v_t$ and heading $h_t$, we construct $\mathcal{G}_t$ by traversing forward until we reach a specified number of intersections (visibility units $u$). An intersection is defined as a node $v$ where $\text{degree}(v) > 2$, indicating a branching point that requires a navigation decision. The algorithm described in Algorithm~\ref{alg:navigate}.
\begin{algorithm}[t]
\small
\caption{Visible Area Construction}
\label{alg:navigate}
\begin{algorithmic}[1]
\REQUIRE node $v_t$, heading $h_t$, visibility units $u$
\ENSURE Path node set $V_{\text{path}}$
\STATE Initialize $V_{\text{path}} \leftarrow \{v_t\}$, $v_{\text{curr}} \leftarrow v_t$, $h_{\text{curr}} \leftarrow h_t$
\STATE $n_{\text{intersections}} \leftarrow 0$
\WHILE{$n_{\text{intersections}} < u$ and $\text{iterations} < 1000$}
    \IF{$\text{degree}(v_{\text{curr}})$ > $2$}
        \STATE $n_{\text{intersections}} \leftarrow n_{\text{intersections}} + 1$
    \ENDIF
    \STATE Get neighbors $N(v_{\text{curr}}) = \{v' \mid (v_{\text{curr}}, v') \in E\}$
    \FOR{each $v' \in N(v_{\text{curr}})$}
        \STATE $h_{v'} \leftarrow \text{bearing}(v_{\text{curr}}, v')$
    \ENDFOR
    \STATE $v_{\text{next}} \leftarrow \arg\min_{v' \in N(v_{\text{curr}})} \Delta h(h_{\text{curr}}, h_{v'})$
    \STATE where $\Delta h(h_1, h_2) = \min(|h_1 - h_2|, 360 - |h_1 - h_2|)$
    \IF{$\Delta h(h_{\text{curr}}, h_{v_{\text{next}}})$ < $100^\circ$}
        \STATE $V_{\text{path}} \leftarrow V_{\text{path}} \cup \{v_{\text{next}}\}$
        \STATE $v_{\text{curr}} \leftarrow v_{\text{next}}$, $h_{\text{curr}} \leftarrow h_{v_{\text{next}}}$
    \ELSE
        \STATE \textbf{break}
    \ENDIF
\ENDWHILE
\STATE Continue for 3 additional nodes for lookahead context
\RETURN $V_{\text{path}}$
\end{algorithmic}
\end{algorithm}

The bearing between two nodes is calculated using the spherical bearing formula (Figure~\ref{fig:overall_architecture}, right):
$$\begin{aligned}
y &= \sin(\Delta \lambda) \cos(\phi_2) \\
x &= \cos(\phi_1) \sin(\phi_2) - \sin(\phi_1) \cos(\phi_2) \cos(\Delta \lambda) \\
h &= \operatorname{atan2}(y, x)
\end{aligned}$$

where $(\phi_1, \lambda_1)$ and $(\phi_2, \lambda_2)$ are the latitude and longitude of the two nodes in radians, and $\Delta \lambda = \lambda_2 - \lambda_1$. The result is normalized to the range $[0, 360)$ degrees.

\paragraph{\ac{POI} Proximity Mapping.}
For each node $v \in V_{\text{path}}$ on the visible path, we identify nearby \acp{POI} within a maximum distance threshold $d_{\max} = 50$ meters. For each grounded landmark $l_m$ with \ac{POI} set $\phi(l_m)$, we calculate the Haversine distance~\cite{inman1849navigation} between each \ac{POI} $p \in \phi(l_m)$ and each path node $v \in V_{\text{path}}$. Then we calculate its relative direction with respect to the current heading. The bearing $h_{v \rightarrow p}$ from node $v$ to \ac{POI} $p$ is calculated using the spherical bearing formula. The relative direction is then determined by the angular difference:
$$\delta = (h_{v \rightarrow p} - h_{\text{curr}} + 180) \bmod 360 - 180$$

The relative direction is classified as follows:
\begin{align*}
\small
\text{direction} =
\begin{cases}
\texttt{Forward} & \text{if } -45^\circ \le \delta \le 45^\circ \\
\texttt{Left}    & \text{if } -135^\circ \le \delta < -45^\circ \\
\texttt{Right}   & \text{if } 45^\circ < \delta \le 135^\circ \\
\texttt{Back}    & \text{otherwise}
\end{cases}
\end{align*}




\paragraph{Action Prediction.}
After receiving the agent's prediction $\text{output}_t$, the system updates the navigation state. If $s_k = \texttt{COMPLETED}$, the current sub-goal index is incremented ($k \leftarrow k + 1$) and the retry counter is reset. If $s_k = \texttt{IN\_PROGRESS}$, the agent moves to the predicted waypoint $v_{t+1}$ while maintaining the same sub-goal. The new heading is calculated as:
$$h_{t+1} = \text{bearing}(v_t, v_{t+1})$$

The navigation process continues until one of the following termination conditions is met: (1) all sub-goals are completed ($k > K$), (2) the maximum total step count is exceeded (100 steps), or (3) the maximum retry count for a single sub-goal is exceeded (15 retries). These threshold values were empirically determined based on the environment characteristics.

\section{Experiment}\label{sec:experiment}
\subsection{Dataset and Implementation Details}

\paragraph{Dataset.}
\begin{table}[ht!]
    \caption{Summary of $\text{TestSet}_A$ and $\text{TestSet}_B$ from the Map2Seq dataset.}
    \label{tab:freq}
    \small
    \centering
    \begin{tabular}{lcc}
    \toprule
    & \textbf{$\text{TestSet}_A$} & \textbf{$\text{TestSet}_B$} \\
    \midrule
        Num. of Instances & 700 & 700 \\
        Avg. Token Length & 53.5 & 54.2  \\
        Landmarks per Instance & 2.72 & 2.69 \\
        Human Nav. Success Rate & 0.86 & 0.84 \\
  \bottomrule
\end{tabular}
\end{table}
We conducted experiments using two test sets from the Map2Seq dataset~\cite{schumann-riezler-2021-map2seq}, which we refer to as $\text{TestSet}_A$ and $\text{TestSet}_B$\footnote{These correspond to the Test\_Seen and Test\_Unseen splits respectively in the original Map2Seq dataset.}. Each dataset contain 700 navigation instances with average instruction lengths of 53.5 and 54.2 tokens respectively. The complete dataset comprises 7,672 crowd-sourced instructions that distinguish themselves from prior work by utilizing a top-down map interface rather than ego-centric panoramic imagery. Annotators were directed to guide a tourist using \ac{OSM} points of interest, which resulted in directions focused on physical objects with an average of 2.7 landmark references per instance. The effectiveness of this approach was confirmed by human navigators who achieved success rates of 0.86 and 0.84 on the $\text{TestSet}_A$ and $\text{TestSet}_B$ respectively within a Street View environment. Table~\ref{tab:freq} summarizes the key statistics for each of these datasets.

\paragraph{Evaluation Metrics.}
Following the evaluation protocols in \cite{liu2023aerialvln}, we validate our method using four core metrics. 
Navigation Error (NE) measures the Euclidean distance between the agent's final position and the ground truth destination. 
Success Rate (SR) indicates the percentage of episodes terminating within 25 meters of the target, while Oracle Success Rate (OSR) is a lenient metric that considers a trial successful if any point on the trajectory falls within this radius.  
Finally, Normalized Dynamic Time Warping (SDTW) evaluates path fidelity by combining binary success status with geometric similarity to the ground truth.
\paragraph{Implementation Details.} 
We implemented our agentic system using the open-source Google Agent Development Kit (ADK)~\footnote{\url{https://google.github.io/adk-docs/}} for real-time evaluation of navigation instructions, alongside standalone scripts to generate and analyze step-by-step requests. These scripts operate in batch mode to evaluate results collectively, which lowers costs.
For \ac{LLM} reasoning, we utilize the online and batch APIs of Gemini, specifically the Gemini-3 Pro model with default parameters (temperature 1.0, and without explicitly defining the thinking level, which defaults to high). Detailed design choices are provided in Appendix~\ref{sec:appendix}, and the full prompts can be found in Appendix~\ref{sec:prompt_templates}.

\begin{table*}[th!]
\caption{Overall navigation execution results; Best-performing baseline methods underlined.}
\label{tab:sir_performance_new}
\centering
\resizebox{0.9\linewidth}{!}{%
\begin{tabular}{l||cccc|cccc}
\toprule
\multirow{3}{*}{\textbf{Method}} 
& \multicolumn{4}{c|}{\textbf{$\text{TestSet}_A$}} 
& \multicolumn{4}{c}{\textbf{$\text{TestSet}_B$}} \\ 
\cmidrule(lr){2-5} \cmidrule(lr){6-9} 
& \textbf{NE $\downarrow$} & \textbf{SR $\uparrow$} & \textbf{OSR $\uparrow$} & \textbf{SDTW $\uparrow$}
& \textbf{NE $\downarrow$} & \textbf{SR $\uparrow$} & \textbf{OSR $\uparrow$} & \textbf{SDTW $\uparrow$}\\
\midrule
Random Walker   & 259.0 & 4.4\% & 5.7\% & 0.026 & 244.3 & 6.1\% & 7.1\% & 0.029 \\
\midrule
Action Sampling & 250.1 & 5.1\% & 6.0\% & 0.037 & 241.6 & 7.4\% & 8.1\% & 0.039 \\
Heuristic Agent & \underline{180.6} & \underline{18.0\%} & \underline{18.9\%} & \underline{0.155} & \underline{173.0} & \underline{17.9\%} & \underline{19.1\%} & \underline{0.159}  \\
\midrule
\textbf{Ours (\approach)} & \textbf{56.8} & \textbf{66.4\%}  & \textbf{78.4\%}  & \textbf{0.634}  & \textbf{59.8}    &    \textbf{63.3\% }    &  \textbf{78.0\%}  &    \textbf{0.609} \\
\bottomrule
\end{tabular}
}
\end{table*}

\subsection{Experimental Results}
\paragraph{Baseline Models.}
To  assess the complexity of the navigation task and quantify the contribution of semantic reasoning of \ac{LLM} agents, we implemented three distinct baseline agents. First, the Random Walker serves as a stochastic baseline to determine the ``chance'' level of success, selecting outgoing edges uniformly at random at every intersection to provide a fundamental lower bound. Second, we employed a rule-based Heuristic Agent to test if navigation is solvable via explicit geometric cues; this agent extracts directional keywords (e.g., \textit{``turn left''}, \textit{``bear right''}) using regular expressions from the navigation instruction and greedily selects the edge best aligned with the command's angle. Finally, the Action Sampling baseline investigates dataset bias by disregarding instruction text entirely and sampling actions based on the pre-computed global probability distribution of ground-truth movements (e.g., the likelihood of moving \textit{``forward''} versus \textit{``left''} at intersections). 

\paragraph{Quantitative Results.}

\begin{table}[ht!]
    \caption{Computational cost statistics comparing average steps and token usage on $\text{TestSet}_A$ and $\text{TestSet}_B$.}
    \small
    \centering
    \label{tab:Computational}
    \begin{tabular}{lcc}
    \toprule
    & \textbf{$\text{TestSet}_A$} & \textbf{$\text{TestSet}_B$} \\
    \midrule
        Avg. Steps & 5.91 & 6.15 \\
        Avg. Thoughts Tokens &  23,044 &  24,000 \\
        Avg. Total Tokens & 44,438 & 46,305 \\
  \bottomrule
\end{tabular}
\end{table}
The experimental results (Table~\ref{tab:sir_performance_new}) demonstrate that our method consistently outperforms 
these fundamental baselines across all metrics on both $\text{TestSet}_A$ and $\text{TestSet}_B$ environments, validating that semantic reasoning provides measurable value beyond chance performance and simple geometric heuristics. Specifically regarding the SR and NE, the proposed method shows substantial superiority over the best-performing baseline Heuristic Agent. In the $\text{TestSet}_A$ the method attains an SR of 66.4\% and an NE of 56.8 while the Heuristic Agent is limited to 18.0\% and 180.6 respectively. A similar performance pattern is evident in the $\text{TestSet}_B$ split. 
These findings validate the premise that structured semantic reasoning provides a viable alternative approach to high-fidelity visual simulations in determining instruction navigability. 
Additionally Table~\ref{tab:Computational} reports the computational cost statistics concerning the average steps and token usage. The results on $\text{TestSet}_A$ and $\text{TestSet}_B$ indicate that the agent takes approximately 5.91 to 6.15 steps on average (median $\tilde{x}=6$ for both). The resource consumption is further detailed with Avg. Total Tokens reaching 44,438 for the $\text{TestSet}_A$ and 46,305 for the $\text{TestSet}_B$.

\paragraph{Correlation Analysis.}
\begin{table}[t]
\caption{Correlation Analysis: Human Annotations vs Navigation Metrics ($n=100$). The metric with the strongest correlation magnitude per column is highlighted in \textbf{bold}. Significance levels are indicated by ** ($p < 0.01$) and * ($p < 0.05$).}
\label{tab:correlation_analysis}
\centering
\resizebox{\linewidth}{!}{%
\begin{tabular}{l||cc|cc}
\toprule
\multirow{2}{*}{\textbf{Metric}} 
& \multicolumn{2}{c|}{\textbf{Pearson Correlation}} 
& \multicolumn{2}{c}{\textbf{Spearman Correlation}} \\
\cmidrule(lr){2-3} \cmidrule(lr){4-5}
& \textbf{$r$} & \textbf{$p$-value} & \textbf{$\rho$} & \textbf{$p$-value} \\
\midrule
SR & 0.2865** & 0.0039 & 0.2865** & 0.0039 \\
OSR & 0.1860 & 0.0639 & 0.1860 & 0.0639 \\
SDTW & 0.2799** & 0.0048 & 0.2860** & 0.0039 \\
nDTW & 0.2457* & 0.0138 & 0.2895** & 0.0035 \\
NE & \textbf{-0.3096**} & 0.0017 & \textbf{-0.3184**} & 0.0012 \\
\bottomrule
\end{tabular}
}
\small\textit{Note:} NE shows a negative correlation, indicating alignment with correct human annotations (lower error = higher score).  \\
\vspace{-8pt}
\end{table}

We randomly sample 100 instructions from Map2Seq Test\_Seen. We present these instructions (along with a static map view) to human annotators and ask them to rate the ``Navigability'' on a binary scale. The human annotators achieved a SR of 86\%, compared to 74\% achieved by the autonomous navigator.
For a more detailed explanation of the navigator's architecture and configuration, refer to Appendix~\ref{sec:appendix}.
Then, we compute the Pearson and Spearman rank coefficients between the human ratings and our metrics (i.e., SR, OSR, NE, SDTW, and nDTW)). 

The results, presented in Table \ref{tab:correlation_analysis}, reveal a statistically significant alignment between human judgment and most automated metrics. Notably, NE demonstrated the strongest correlation with human annotations across both Pearson ($r = -0.31$, $p < 0.01$) and Spearman ($\rho = -0.32$, $p < 0.01$) coefficients. The negative correlation confirms that lower navigation errors consistently correspond to higher human ratings of trajectory quality. Furthermore, while SR and SDTW showed moderate positive correlations ($r \approx 0.29$), OSR failed to achieve statistical significance ($p > 0.05$), suggesting it may be a less reliable proxy for human-perceived navigation quality in this context. These findings support the use of NE and nDTW as primary metrics for evaluating agent performance when human-in-the-loop validation is not feasible.

\section{Conclusions \& Future work}\label{sec:final_remarks}
In this paper, we introduced \approach, a vision-free framework for evaluating navigation instructions using \ac{OSM} data. We proposed an inversion of the standard \ac{VLN} task where we treat agent execution success as a metric for instruction quality rather than agent capability. Our ablation studies revealed that structured JSON representations combined with hierarchical sub-instruction planning enable \acp{LLM} to reason effectively about spatial graphs. Specifically, we demonstrate that the \acp{LLM} interpret instructions better as sub-goals, allowing agents to focus on simple actions and environmental reasoning.
This approach yields performance metrics that correlate significantly with human judgments of navigability and eliminates the noise introduced by visual perception failures in traditional simulators.

For future work, we aim to reduce computational overhead by training domain-specific small language models. We plan to use teacher-student distillation to create compact models suitable for edge deployment. This optimization will facilitate the integration of our system with assistive technologies and smart devices. Specifically, we intend to explore how wearable sensors and smart glasses can capture environmental data to aid human navigation in real-time. By processing visual or auditory inputs into topological graph data, these devices can provide accessible guidance grounded in the evaluated instructions.

\section*{Limitations}
Our vision-free \approach restricts the scope of evaluation to structural and semantic navigability. The agent operates solely on symbolic representations and therefore cannot validate instructions that rely heavily on purely visual cues not encoded in the map schema, such as \textit{``turn left at the house with the red door''} or \textit{``follow the graffiti wall''}. As a result, the framework is less suitable for evaluating instruction-following in environments where visual grounding is the dominant mode of guidance. At the same time, this abstraction enables controlled evaluation of safety-critical and cognitively relevant aspects of navigation that are often overlooked in vision-centric setups. In particular, the framework is well-suited for integration with assistive navigation systems, such as smart glasses or wearable guidance devices, where symbolic reasoning, instruction clarity, and safety constraints play a central role. This allows the evaluation of navigational reliability and ambiguity in scenarios that cannot be easily captured through visual perception alone.

Regarding computational efficiency, the reliance on high-reasoning LLMs introduces significant latency and cost. Our efficiency analysis indicates that achieving optimal trajectory fidelity requires the ``High'' thinking configuration, which consumes approximately 41,347 tokens per average episode. While this overhead limits the \approach’s suitability for large-scale deployment, the resulting execution traces and decision trajectories provide a valuable resource for case-study-driven analysis and targeted model training. In particular, the collected episodes can support the development and fine-tuning of models with improved graph comprehension and instruction execution capabilities, enabling future systems to approximate similar performance with substantially reduced inference costs.

Finally, the empirical validity of our findings regarding spatial representations is currently limited to the Gemini-3 Pro architecture. While we demonstrated that structured JSON outperforms grid-based formats for this specific model, we have not yet established whether this preference for hierarchical data is a universal characteristic of all \acp{LLM} or an artifact of the specific training distribution of the model employed. Future work must verify these findings across a broader spectrum of proprietary and open-weights models to ensure generalizability.

\bibliography{custom}

\appendix

\section{Ablation Studies}\label{sec:appendix}
We conducted ablation studies to assess the core components of the proposed method. We randomly sampled 100 instances from the seen validation dataset of Map2Seq and performed all the ablation experiments. We analyzed different types of spatial information presentation in Section~\ref{sec:spatial_info}, which corresponds to the navigator agent. Then, in Section~\ref{sec:sub_instructions}, we assessed the effectiveness of the Sub-instruction divider agent.  In Section~\ref{sec:poi_detection}, we evaluated the accuracy of \ac{POI} detection using different models compared to human-level performance. Finally, in Section~\ref{sec:thinking_level}, we examined how the thinking procedure in these models affects the final results to determine if LLM reasoning aids the process and justifies the computational cost. Additionally, we studied the relationship between instructions and how sub-instruction updates facilitate the navigation process.

\paragraph{Data annotation.} 
To ensure better data interpretability, we employ human annotators to process the navigation instructions. Specifically, we validate instruction difficulty and navigability, while also annotating \acp{POI} mentions within the text.
For difficulty assessment, we independently evaluated and rated each navigation instruction on a scale of 1 to 10 across three distinct dimensions which we have defined as Linguistic Intricacy, Topological Complexity, and Operational Demand.

We calculated the average rating for each instruction and normalized the scores to classify them into three subcategories. Terminology and descriptions for these measurements are defined below:
\begin{itemize}
    \item \textbf{Linguistic Intricacy:} This metric assesses the semantic load required to interpret the instruction. We evaluate the visibility of attributes alongside the density of the text including the frequency of action verbs and the specificity of landmark references.
    \item \textbf{Topological Complexity:} This metric pertains to the graph network and the physical layout of the environment. We consider factors such as the total Euclidean distance, the spatial coverage of the area, and the geometry of the path within the map structure.
    \item \textbf{Operational Demand:} This metric concentrates on the physical effort required to traverse the path. We measure this by analyzing the quantity of discrete steps, the number of required turns, and the frequency of transitions between different areas or floors.
\end{itemize}

\begin{figure}[t]
  \includegraphics[width=\linewidth]{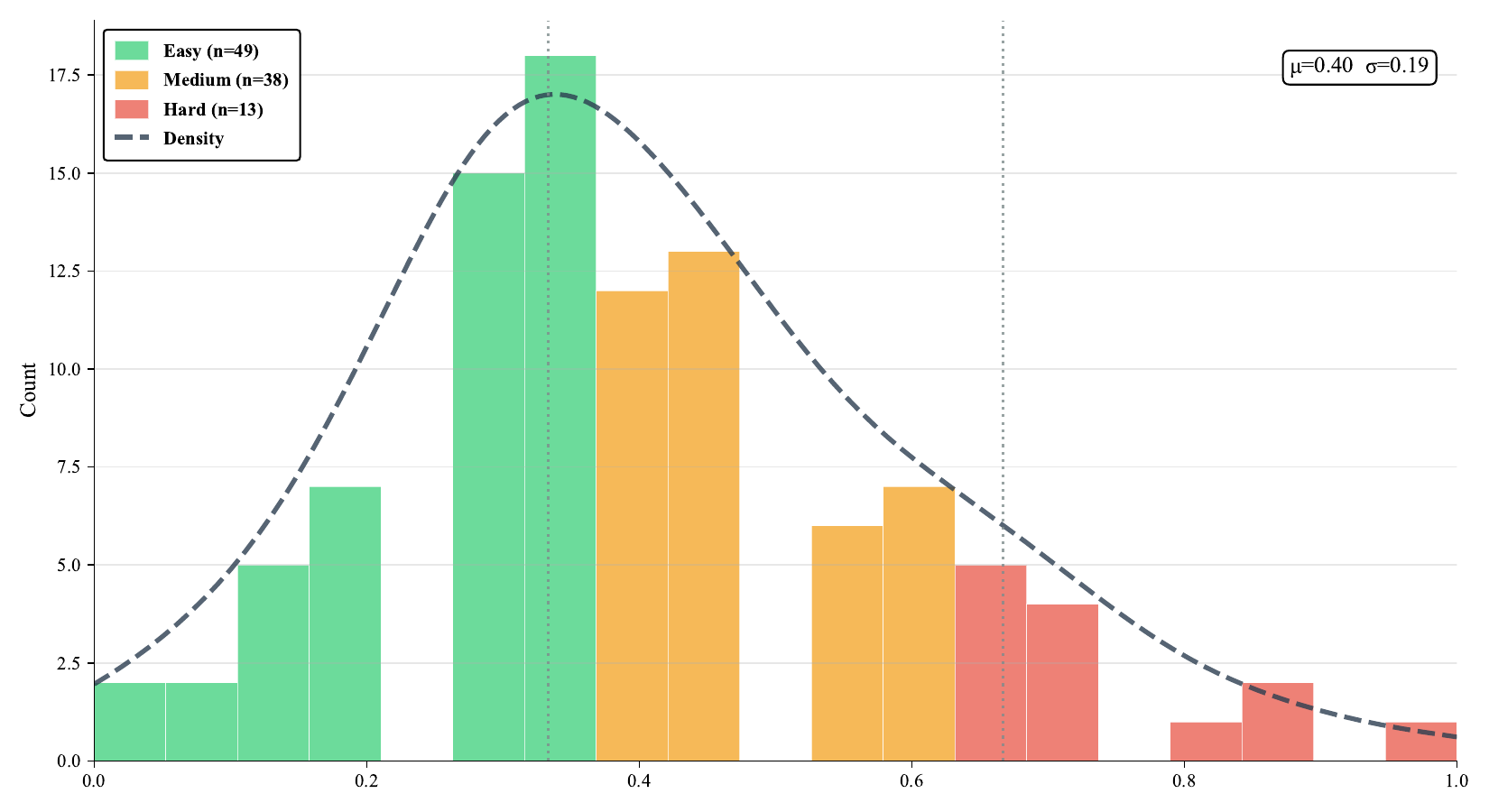}
  \caption{Normalized complexity score distribution.}
  \label{fig:complexity_score_distribution}
\end{figure}

As result, we categorized each case into one of three levels: (1) Easy where cases consist of short paths accompanied by simple instructions that require minimal spatial reasoning and utilize clear landmarks; (2) Medium where cases involve instructions of moderate length that contain multiple spatial phrases and dictate traversal over medium-length paths; and (3) Hard in which the cases involve long trajectories guided by complex multi-step instructions which often necessitate floor transitions and the processing of multiple spatial references. Figure \ref{fig:complexity_score_distribution} illustrates the normalized complexity score distribution following the annotation process. The dataset is categorized into 49 easy, 38 medium, and 13 hard instructions. Statistically, the distribution yields a mean of $\mu=0.40$ and a standard deviation of $\sigma=0.19$.

For the navigability assessment, we defined correctness as the human evaluator arriving within a 25-meter radius of the designated end location in binary scale (0-1).

Finally, we manually tag and count \acp{POI} mentioned in the navigation instructions. When the same \ac{POI} type appears multiple times within a single instruction, we count it as one instance. For example, if an instruction contains several references to traffic lights (such as \textit{``turn at the traffic light''} and \textit{``pass the traffic light''}), we record this as a single \ac{POI}.

\subsection{Different Types of Spatial Information.}\label{sec:spatial_info}
To demonstrate how distinct spatial representations influence LLM performance, we compared various prompting formats for encoding spatial information. We evaluated four core representation strategies for graph and spatial data: (1) Textual or incident encoding; (2) Structured JSON format; (3) Graphviz-style visual representation; and (4) Matrix or grid representation. These design choices take inspiration from the works by~\citet{gao2024aerial,fatemi2024talk} and ~\citet{zhao2024graphtext}.

In textual encoding, we characterize the graph structure by enumerating each node's direct connections. This approach represents both nodes and their connectivity through natural language text attributes. For each node, we describe its connections to adjacent nodes with heading information in degrees. 

\begin{tcolorbox}[
    enhanced,
    colback=prompt1_bg,
    colframe=prompt1_frame,
    boxrule=2pt,
    arc=4mm,
    width=\linewidth,
    title={Textual Format Example},
    coltitle=white,
    fonttitle=\bfseries,
    attach boxed title to top left={yshift=-3mm, xshift=6mm},
    boxed title style={colback=prompt1_frame, arc=3mm},
    breakable,
    fontupper=\small
]
    \vspace{0.2cm}
     
    Node 38eb:\\
    \hspace*{1em}Connected to nodes:\\
    \hspace*{2em} - Node 4242 is to the forward (heading: 208.6$^\circ$, Southwest)\\

    Intersection 4242:\\
    \hspace*{1em} Connected to nodes:\\
    \hspace*{2em}- Node cbc2 is to the forward (heading: 208.9$^\circ$, Southwest)\\
    \hspace*{2em}- Node 5b89 is to the left (heading: 119.0$^\circ$, Southeast)\\
    \hspace*{1em}Branches from this intersection:\\
    \hspace*{2em}- Forward branch (heading: 208.9$^\circ$, Southwest):\\
    \hspace*{3em}- Path: cbcf $\rightarrow$ c5d0\\
    ....
\end{tcolorbox}

The textual format includes \ac{POI} legends, current position markers, compass directions, and intersection branch exploration where applicable. Each node description contains its type (intersection or waypoint), connections with headings $h \in [0, 360)$, and nearby \acp{POI} with directional and distance information.

For the Structured JSON format, we maintain the spatial data hierarchically in a machine-readable structure. The JSON representation organizes nodes and \acp{POI} into separate sections. Each node entry contains its identifier, type classification, heading value $h$, connection list with target node IDs and headings, and optional coordinate data $(\textit{lat}, \textit{lng})$ when available. Intersection nodes include branch information with extended node sequences per direction (Left, Right, Forward). The \ac{POI} section lists all landmarks with their assigned letters, nearby node references, directional information, and distance measurements in meters. This hierarchical organization facilitates structured parsing and programmatic access to spatial relationships.

Regarding visual language representations, the first approach uses Graphviz-style notation that explicitly delineates nodes and edges using arrow syntax. For example, \texttt{13480 $\rightarrow$ 78640 [heading: $30^\circ$, direction: Forward]}. 
By separating node definition from edge definition with clear arrow notation, it provides a significantly cleaner structure for the LLM to parse. The Graphviz format supports both arrow-style notation and full DOT format syntax. It includes intersection markers, branch chains with multiple nodes per direction, and \ac{POI} connections marked with dashed edges and distance annotations.

\begin{tcolorbox}[
    enhanced,
    colback=prompt1_bg,
    colframe=prompt1_frame,
    boxrule=2pt,
    arc=4mm,
    width=\linewidth,
    title={Graphvis Style Format Example},
    coltitle=white,
    fonttitle=\bfseries,
    attach boxed title to top left={yshift=-3mm, xshift=6mm},
    boxed title style={colback=prompt1_frame, arc=3mm},
    breakable,
]
    \noindent\resizebox{\linewidth}{!}{%
    \begin{minipage}{\linewidth} 
        \begin{align*} 
        &\texttt{5fa6} \rightarrow \texttt{946a [heading: 208$^\circ$, direction: Forward]} \\
        &\texttt{946a} \rightarrow \texttt{ec11 [heading: 208$^\circ$, direction: Forward]} \\
        &\texttt{ec11} \rightarrow \texttt{4242 [heading: 208$^\circ$, direction: Forward]}
        \\\\
        &\texttt{4242[Intersection]} \rightarrow \texttt{5c7f [heading: 208$^\circ$, direction: Forward]} \\
        &\texttt{4242[Intersection]} \rightarrow \texttt{501f [heading: 208$^\circ$, direction: Forward]} \\
        &\texttt{4242[Intersection]} \rightarrow \texttt{db40 [heading: 208$^\circ$, direction: Forward]} \\\\
        &\textbf{Intersection Branches (extended nodes)} \\
        &\texttt{4242} \xrightarrow{\text{Forward}} \texttt{5c7f} \rightarrow \texttt{d4cc [heading: 208$^\circ$, Southwest]} \\
        &\texttt{4242} \xrightarrow{\text{Left}} \texttt{501f} \rightarrow \texttt{bf02 [heading: 119$^\circ$, Southeast]} \\
        &\texttt{4242} \xrightarrow{\text{Right}} \texttt{db40} \rightarrow \texttt{84bc [heading: 298$^\circ$, Northwest]} \\
    \end{align*}
    \end{minipage}%
}
\end{tcolorbox}

The second visual approach positions \ac{OSM} nodes within two-dimensional grids to render the map in a pixel-wise text format. This representation converts the spatial graph into a matrix $G \in \mathbb{R}^{H \times W}$ where each cell $G[i,j]$ contains a character representing the spatial state. We employ \ac{BFS} traversal starting from a start node to establish relative coordinates for the grid construction. The position of each node relies on heading-based directional offsets where we map headings $h$ to directional vectors $(dr, dc)$. Headings where $315^\circ \leq h < 360^\circ$ or $0^\circ \leq h < 45^\circ$ map to North $(-1, 0)$, $45^\circ \leq h < 135^\circ$ map to East $(0, 1)$, $135^\circ \leq h < 225^\circ$ map to South $(1, 0)$, and $225^\circ \leq h < 315^\circ$ map to West $(0, -1)$.

We design a logic for \ac{POI} integration to maintain spatial accuracy without cluttering the path. We first assess if a \ac{POI} resides within a defined threshold distance, typically 20 meters, of an intersection node. If the \ac{POI} is near an intersection, we calculate the heading from the intersection to the \ac{POI} to determine the correct diagonal quadrant. We place these corner \acp{POI} at diagonal offsets, such as $(-1, 1)$ for Northeast or $(1, 1)$ for Southeast, which ensures they occupy cells that do not intersect with the cardinal route segments. For \acp{POI} located along standard path segments, we calculate the heading from the nearest path node and position the \ac{POI} in the adjacent cell corresponding to that direction. If a \ac{POI} is extremely close to the node, typically under 5 meters, it may share the grid position unless that position is reserved for start or current markers.

\begin{figure}
    \centering
    \includegraphics[width=\linewidth]{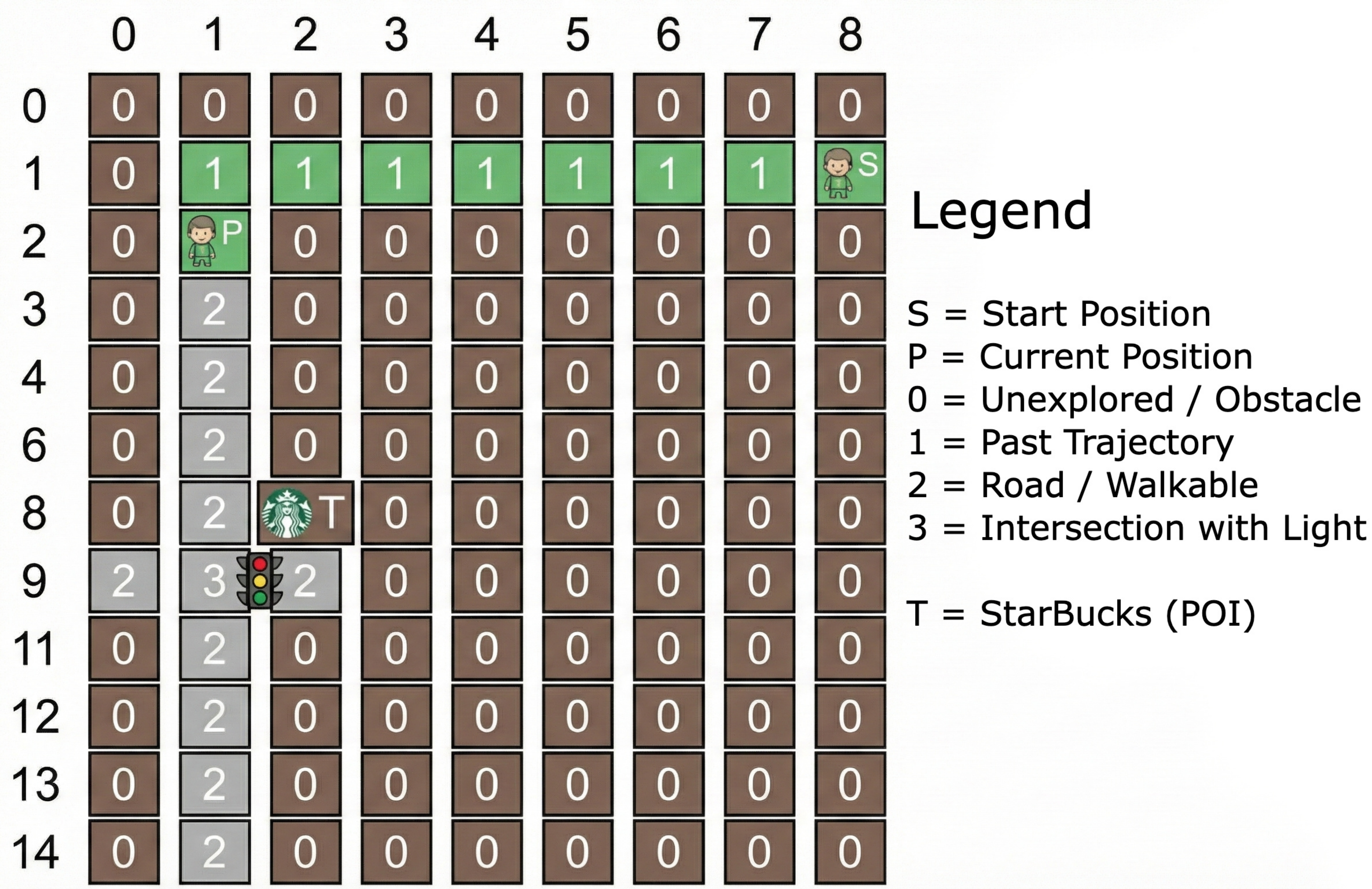}
    \caption{Matrix / grid representation for the second iteration of the instruction: \textit{``Make two left turns to head the opposite way on the other side of the block.''} The first left turn is already executed, as indicated by the past trajectory denoted as `1' in the representation.}
    \label{fig:grid-repsentation}
\end{figure}

The grid utilizes a specific character set to denote state configuration and facilitate LLM interpretation. `S' indicates the initial starting position and `P' marks the current agent location. We use `1' to represent previously visited nodes and `2' to denote the current forward path segments. Intersections are explicitly marked with `3', while `0' denotes empty space. We represent specific landmarks using single characters defined in a dynamically created ``poi\_mapping'' dictionary, and thereby creating a dynamic legend where a landmark name maps to a specific letter. This mapping is either provided explicitly or extracted dynamically from the path data to ensure that every visible \ac{POI} has a unique identifier within the grid context. Figure~\ref{fig:grid-repsentation} shows an example of the grid presentation for specific navigation instruction.

For all evaluation modes, we constrained the input data to mitigate hallucinations within the \ac{LLM}. We acknowledge that excessive context size increases computational overhead and the probability of information retrieval errors. As our objective is to develop a navigation agent, we exclusively provide the immediate surroundings, effectively simulating the limited field of view visible to a human observer. This constraint limits the number of nodes and \acp{POI} included in each representation while preserving the essential spatial connectivity information required for navigation decision-making.

\begin{table*}[th!]
\caption{Performance comparison on Navigation Execution for different types of spatial information representation categorized by the instruction difficulty. The bottom row reports optimized representation’s improvement over the best existing method.}
\vspace{-2pt}
\label{tab:sir_performace}
\centering
\resizebox{\linewidth}{!}{%
\begin{tabular}{l||cccc|cccc|cccc}
\toprule
\multirow{3}{*}{\shortstack{\textbf{Thinking} \\ \textbf{Level}}} 
& \multicolumn{12}{c}{\textbf{Navigation Execution}} \\ 
\cmidrule(lr){2-13} 
& \multicolumn{4}{c|}{\textbf{Easy}} 
& \multicolumn{4}{c|}{\textbf{Medium}} 
& \multicolumn{4}{c}{\textbf{Hard}} \\
\cmidrule(lr){2-5} \cmidrule(lr){6-9} \cmidrule(lr){10-13} 
& \textbf{NE $\downarrow$} & \textbf{SR$\uparrow$} & \textbf{OSR$\uparrow$} & \textbf{nDTW$\uparrow$} 
& \textbf{NE $\downarrow$} & \textbf{SR$\uparrow$} & \textbf{OSR$\uparrow$} & \textbf{nDTW$\uparrow$} 
& \textbf{NE $\downarrow$} & \textbf{SR$\uparrow$} & \textbf{OSR$\uparrow$} & \textbf{nDTW$\uparrow$} \\
\midrule
Textual & \colorbox{purple_color!7}{\textbf{71.3}} & \colorbox{purple_color!25}{\textbf{61.2\%}} & \colorbox{purple_color!7}{\textbf{69.4\%}} & \colorbox{purple_color!7}{\textbf{0.602}} & \colorbox{purple_color!25}{\textbf{56.6}} & \colorbox{purple_color!25}{\textbf{68.4\%}} & \colorbox{purple_color!7}{\textbf{71.1\%}} & \colorbox{purple_color!25}{\textbf{0.637}} & \colorbox{purple_color!25}{\textbf{110.6}} & \colorbox{purple_color!7}{\textbf{38.5\%}} & \colorbox{purple_color!7}{\textbf{46.2\%}} & \colorbox{purple_color!7}{\textbf{0.380}} \\
JSON & \colorbox{purple_color!25}{\textbf{62.1}} & \colorbox{purple_color!25}{\textbf{61.2\%}} & \colorbox{purple_color!25}{\textbf{75.5\%}} & \colorbox{purple_color!25}{\textbf{0.603}} & \colorbox{purple_color!7}{\textbf{61.2}} & \colorbox{purple_color!25}{\textbf{68.4\%}} & \colorbox{purple_color!25}{\textbf{78.9\%}} & \colorbox{purple_color!25}{\textbf{0.637}} & \colorbox{purple_color!7}{\textbf{112.9}} & \colorbox{purple_color!25}{\textbf{53.8\%}} & \colorbox{purple_color!25}{\textbf{53.8\%}} & \colorbox{purple_color!25}{\textbf{0.525}} \\
Graphvis-style & 90.4 & \colorbox{purple_color!7}{\textbf{40.8\%}} & 53.1\% & 0.401 & 87.8 & \colorbox{purple_color!7}{\textbf{47.4\%}} & 52.6\% & \colorbox{purple_color!7}{\textbf{0.453}} & 146.5 & 15.4\% & 30.8\% & 0.151 \\
Grid & 186.7 & 6.1\% & 8.2\% & 0.060 & 160.3 & 13.2\% & 13.2\% & 0.128 & 176.6 & 15.4\% & 23.1\% & 0.141 \\
\midrule
\textbf{Optimized Repr.} & \textbf{35.6} & \textbf{77.6\%} & \textbf{85.7\%} & \textbf{0.762} & \textbf{30.9} & \textbf{76.3\%} & \textbf{84.2\%} & \textbf{0.715} & \textbf{93.3} & \textbf{53.8\%} & \textbf{61.5\%} & \textbf{0.527} \\
\midrule
\textbf{Improve.} & 42.7\% & 26.8\% & 13.5\% & 26.4\% & 45.4\% & 11.5\% & 6.7\% & 12.2\% & 15.6\% & 0.0\% & 14.3\% & 0.4\% \\
\bottomrule
\end{tabular}
}
\small\textit{Note:} \colorbox{purple_color!25}{\textbf{Dark purple}} and \colorbox{purple_color!7}{\textbf{light purple}} indicate the top-performing and second top-performer per column. \\
\vspace{-8pt}
\end{table*}

\begin{figure*}
\centering
    \includegraphics[width=\linewidth]{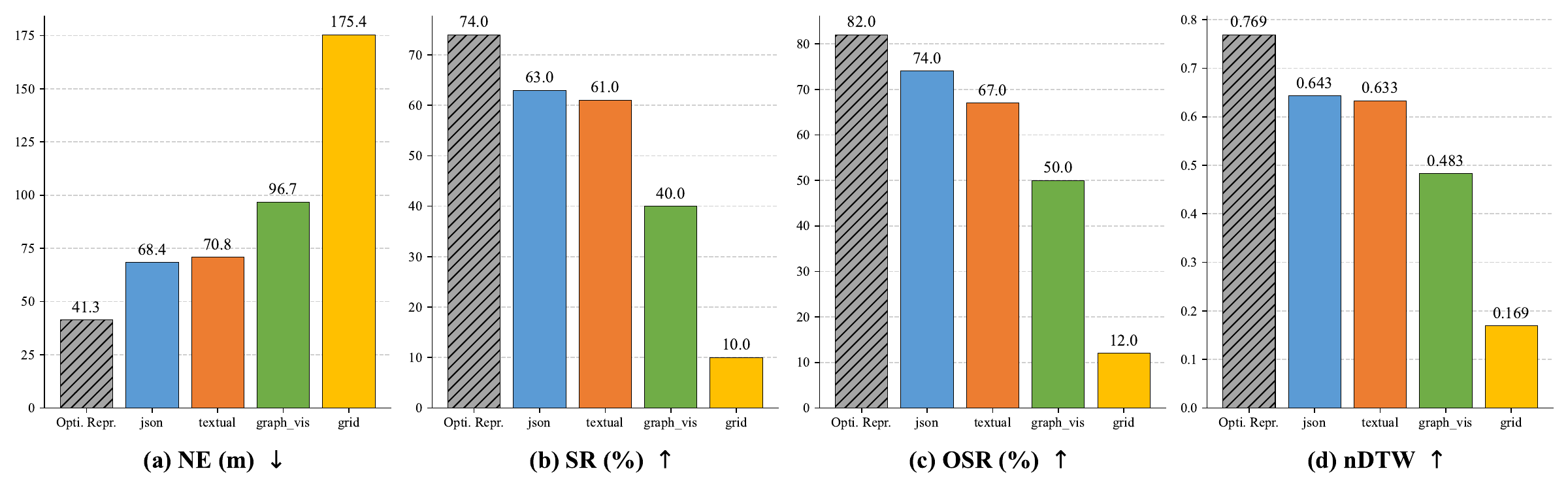}
    \caption{Overall performance comparison on Navigation Execution for different types of spatial information representation.}
    \label{fig:sir_performace}    
\end{figure*}

\paragraph{Results.} Figure~\ref{fig:sir_performace} and Table~\ref{tab:sir_performace} summarize the quantitative results  for different types of spatial information representation.

As illustrated in Figure~\ref{fig:sir_performace}, the Structured JSON and Textual encodings significantly outperform the visual language approaches. The JSON format achieves the most favorable balance across metrics, recording a Navigation Error of 68.4 meters and a Success Rate of 63.0\%. The Textual representation follows closely with an NE of 70.8 meters and an SR of 61.0\%. While their navigation success rates are comparable, the JSON format demonstrates superior path fidelity. It achieves a higher nDTW score of 0.643 compared to 0.633 for the Textual format and a notably higher Oracle Success Rate of 74.0\% versus 67.0\%. This indicates that the hierarchical structure of JSON may allow the agent to recover from deviations more effectively than the linear narrative of the Textual format.

The visual encoding strategies lag behind the semantic formats. The Graphviz-style representation results in a significantly higher NE of 96.7 meters and a reduced SR of 40.0\%. The Grid representation proves to be the least effective method for this task. It yields the highest global Navigation Error of 175.4 meters and a minimal Success Rate of 10.0\%. The extremely low nDTW of 0.169 for the Grid format suggests that the LLM struggles to parse high-density ASCII matrices or derive spatial relationships from pixel-wise text layouts. The predominant error in this mode involves the selection of `0' cells representing unexplored areas, even though the instructions explicitly prohibited such actions. 

Table~\ref{tab:sir_performace} reveals how performance diverges as instruction complexity increases. In the \textbf{Easy} and \textbf{Medium} categories, the distinction between Textual and JSON formats is minimal. For Medium difficulty tasks, both formats achieve an identical Success Rate of 68.4\%. The Textual format even records a slightly better Navigation Error in this specific tier (56.6m vs 61.2m).

However, in the \textbf{Hard} category, the performance of the Textual representation degrades sharply, with the Success Rate dropping to 38.5\%. In contrast, the JSON format maintains a significantly higher SR of 53.8\% and a higher nDTW score of 0.525 compared to 0.380 for Textual. This divergence suggests that while natural language descriptions are sufficient for simple routing, the cognitive load associated with parsing complex, multi-step textual instructions hampers reasoning. 

\begin{table*}[th!]
    \centering
    \caption{Error analysis of wrongly predicated path for navigation instructions in JSON presentation.}
    \label{tab:error_analysis}
    \resizebox{\linewidth}{!}{
    \begin{tabular}{llc}
    \toprule
    \textbf{Error Category} & \textbf{Description} & \textbf{Frequency (Count)} \\
    \midrule
    Spatial Grounding & Failure to detect \acp{POI} or map semantic labels to inputs. & 17  \\
    Spatial Enumeration & Errors in counting discrete landmarks (e.g., \textit{``3rd light"}). & 6  \\
    State Estimation & Failures in heading initialization or instruction pointer updates (re-runs). & 5  \\
    Ambiguity / Dataset Error & Vague instructions or incorrect ground truth labels. &  5  \\
    Topological Mismatch &  Agent fails at node due to descriptions and environmental layout mismatch. & 4  \\
    \bottomrule
    \end{tabular}}
\end{table*}

\paragraph{Optimized Representation.} Since the Structured JSON format appears to be the most promising, we analyzed all the failure cases.  As shown in Table~\ref{tab:error_analysis}, the errors in JSON presentations relate primarily to Spatial Grounding, specifically the model's failure to detect \acp{POI} and map semantic relationships, and Spatial Enumeration Errors involving the counting of discrete landmarks. These two types of error contribute to 62\% of all errors in this category. In addition, there are cases where the Agent traverses the correct path but fails to stop at the specific node. 

To address mentioned issues in JSON presentations, we added more criteria to the prompt and included clarifications for handling \acp{POI}. Additionally, we appended the iteration number to the \texttt{IN\_PROGRESS} sub-instruction to assist the \acp{LLM} in tracking counts, such as when passing the ``3rd light.'' We also observed that providing details like \textit{lat} and \textit{lng} complicated the decision process rather than improved the navigation ability, so we removed these coordinates from the JSON prompts. The final prompt for the navigator agent is provided in Prompt~\ref{prompt:nav_agent}.

The last row of Table~\ref{tab:sir_performace} and the hatched bar in Figure~\ref{fig:sir_performace} display the results of the modified JSON representation regarding this error analysis, showing substantial relative improvements of up to 45.4\% in navigation error reduction and 26.8\% in success rate gains compared to the best baseline methods. Consequently, we exclusively utilize this optimized representation for all future experiments.

\begin{figure}
    \centering
    \includegraphics[width=\linewidth]{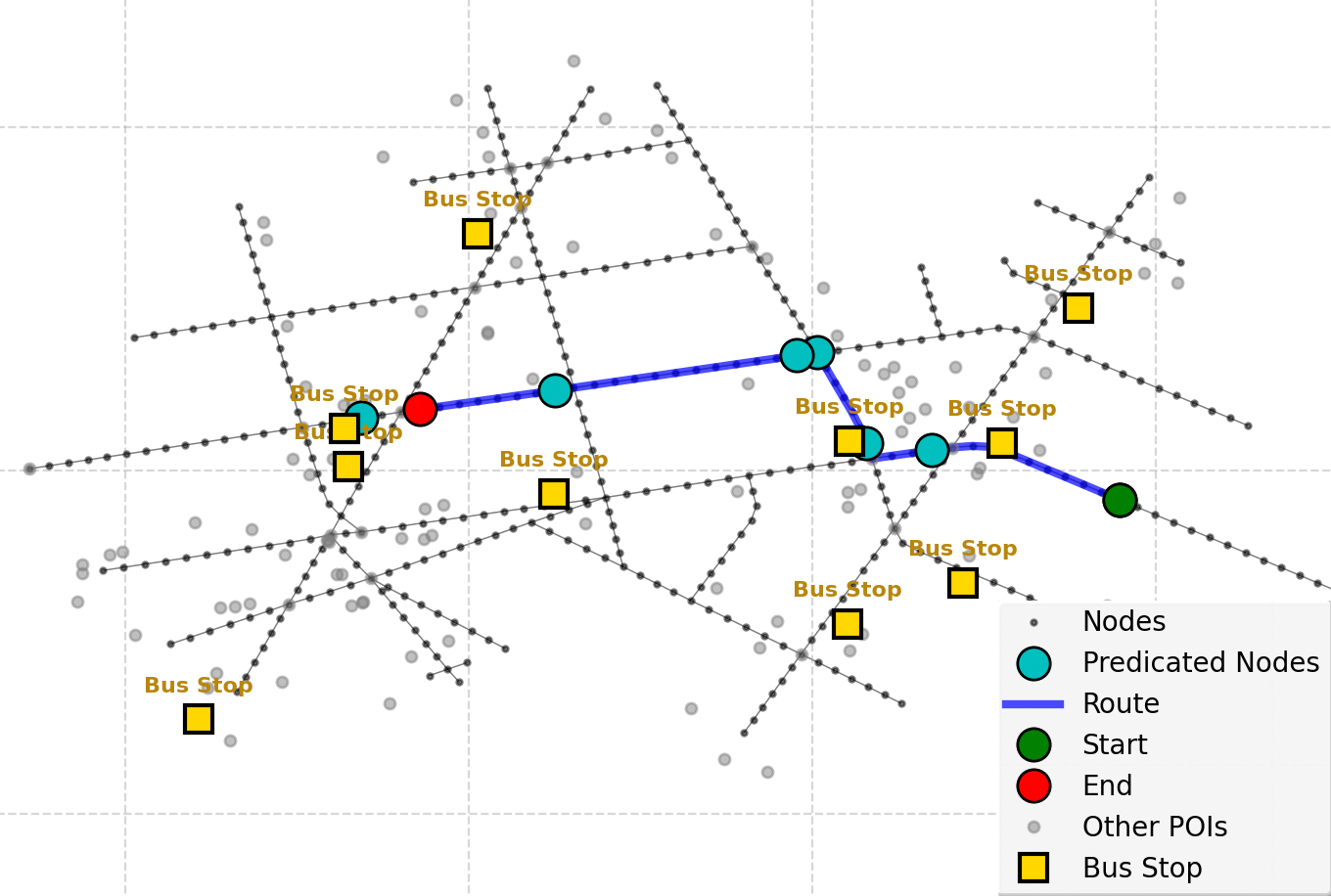}
    \caption{Visualization of a navigation failure where the agent overshoots the intended goal. The map displays the ground truth path and the agent's route, highlighting how the agent incorrectly interprets the bus stop landmark as the destination rather than a reference point for the stopping location.}
    \label{fig:case_study}
\end{figure}

\paragraph{A Case Study.} Figure~\ref{fig:case_study} illustrates one of the most frequent failure cases caused by incorrect planning or execution. In this scenario, the navigation instruction is as follows: \textit{``Go straight through 1 light and at the following light very soon after the 1st, a garden or square should be on the near right corner. Turn right and walk to the next light. Turn left and pass a 4-way intersection and stop just before entering the traffic light. A bus stop should be on the far left corner.''} As demonstrated in the figure, the objective is to stop immediately before the intersection. However, the agent extends the path further because it misinterprets the intention of the guide as providing an address to reach the bus stop itself. Generally, we can conclude that this specific typex of failure originates from the misunderstanding of ambiguous instructions regarding the final stopping condition.

\subsection{Sub-instructions \& Planning}\label{sec:sub_instructions}
Here, we investigated two distinct aspects concerning sub-instruction, where we first evaluated the importance of the existence of a divider agent and subsequently analyzed how planning and state management facilitate the \acp{LLM} in the execution of instructions.

\paragraph{Sub-instruction Divider Agent.} To test the impact of dividing the instruction into sub-instructions using \acp{LLM} for better interpretability, we conducted three different tests. In the first method, we prompted the \acp{LLM} with the complete instruction, whereas in the second method, we employed a simple rule-based approach where the text was split using periods. Finally, we provided the sub-instructions produced by the same \ac{LLM} as steps to the \ac{LLM}. Although we agree that \acp{LLM} are capable of performing tasks, achieving reasonable results requires defining the task in a simple and clear manner. As the results demonstrate, the model can interpret the instruction better when instructions are defined as sub-goals. Furthermore, the model can focus more effectively and reason better about the surrounding environment when instructions are divided into simple actions, such as moving forward or turning, combined with specific criteria like \textit{``after X''} or \textit{``when seeing Y.''}

\begin{table*}[th!]
\caption{Performance comparison on Navigation Execution for different sub-instruction methods categorized by the instruction difficulty. The bottom row reports the LLM Divider's improvement over the best baseline method.}
\vspace{-2pt}
\label{tab:subi_performace}
\centering
\resizebox{\linewidth}{!}{%
\begin{tabular}{l||cccc|cccc|cccc}
\toprule
\multirow{3}{*}{\textbf{Method}} 
& \multicolumn{12}{c}{\textbf{Navigation Execution}} \\ 
\cmidrule(lr){2-13} 
& \multicolumn{4}{c|}{\textbf{Easy}} 
& \multicolumn{4}{c|}{\textbf{Medium}} 
& \multicolumn{4}{c}{\textbf{Hard}} \\
\cmidrule(lr){2-5} \cmidrule(lr){6-9} \cmidrule(lr){10-13} 
& \textbf{NE $\downarrow$} & \textbf{SR$\uparrow$} & \textbf{OSR$\uparrow$} & \textbf{nDTW$\uparrow$} 
& \textbf{NE $\downarrow$} & \textbf{SR$\uparrow$} & \textbf{OSR$\uparrow$} & \textbf{nDTW$\uparrow$} 
& \textbf{NE $\downarrow$} & \textbf{SR$\uparrow$} & \textbf{OSR$\uparrow$} & \textbf{nDTW$\uparrow$} \\
\midrule
Complete Instr. & \colorbox{purple_color!7}{\textbf{49.6}} & \colorbox{purple_color!7}{\textbf{59.2\%}} & 65.3\% & \colorbox{purple_color!7}{\textbf{0.581}} & 71.3 & 50.0\% & 52.6\% & 0.477 & 157.9 & 23.1\% & 23.1\% & 0.229 \\
Rule-based Split & 58.9 & 49.0\% & \colorbox{purple_color!7}{\textbf{69.4\%}} & 0.478 & \colorbox{purple_color!7}{\textbf{40.7}} & \colorbox{purple_color!7}{\textbf{65.8\%}} & \colorbox{purple_color!7}{\textbf{78.9\%}} & \colorbox{purple_color!7}{\textbf{0.592}} & \colorbox{purple_color!7}{\textbf{108.2}} & \colorbox{purple_color!7}{\textbf{38.5\%}} & \colorbox{purple_color!7}{\textbf{38.5\%}} & \colorbox{purple_color!7}{\textbf{0.382}} \\
\midrule
\textbf{LLM Divider} & \colorbox{purple_color!25}{\textbf{35.6}} & \colorbox{purple_color!25}{\textbf{77.6\%}} & \colorbox{purple_color!25}{\textbf{85.7\%}} & \colorbox{purple_color!25}{\textbf{0.762}} & \colorbox{purple_color!25}{\textbf{30.9}} & \colorbox{purple_color!25}{\textbf{76.3\%}} & \colorbox{purple_color!25}{\textbf{84.2\%}} & \colorbox{purple_color!25}{\textbf{0.715}} & \colorbox{purple_color!25}{\textbf{93.3}} & \colorbox{purple_color!25}{\textbf{53.8\%}} & \colorbox{purple_color!25}{\textbf{61.5\%}} & \colorbox{purple_color!25}{\textbf{0.527}} \\
\midrule
\textbf{Improve.} & 28.2\% & 31.1\% & 23.5\% & 31.2\% & 24.1\% & 16.0\% & 6.7\% & 20.8\% & 13.8\% & 39.7\% & 59.7\% & 38.0\% \\
\bottomrule
\end{tabular}
}
\small\textit{Note:} \colorbox{purple_color!25}{\textbf{Dark purple}} and \colorbox{purple_color!7}{\textbf{light purple}} indicate the top-performing and second top-performer per column. \\
\vspace{-8pt}
\end{table*}

Table~\ref{tab:subi_performace} summarizes the results across these three methodologies. Based on the results obtained, the data indicates that breaking down complex directives significantly reduces ambiguity and improves the execution success rate. 

Notably, the \textbf{LLM Divider} method consistently surpasses the rule-based splitting, suggesting that intelligent, context-aware segmentation is better than rigid syntactic splitting based on punctuation. This advantage is particularly pronounced in the ``Hard'' difficulty settings, where the divider agent maintains a higher performance with a 53.8\% success rate, whereas the baselines degrade significantly. Ultimately, transforming instructions into explicit sub-goals allows the agent to reason more effectively about immediate actions without losing track of the global objective.

\paragraph{Planning State.} In addition, we manually inspected the impact of planning regarding the state update of sub-instructions and their dependency. The first insight reveals that although \acp{LLM} can perform complex tasks, the model encounters difficulties when a sub-instruction requires repeating a movement multiple times, such as \textit{``Continue straight through the next three intersections.''} In these scenarios, even though we provided the previous steps the agent had already passed, we must specifically define the current iteration count for the sub-instruction in progress to avoid errors related to repeated movements. Furthermore, certain instructions require waiting until a specific condition is satisfied, such as moving to the end of next block to ensure the visibility of \acp{POI} mentioned in the text. Consequently, it is impossible to track the path effectively if we do not maintain states for each sub-instruction. The final case we identified involves human-generated navigation instructions where actions in different instructions are relevant to each other, such as \textit{``move forward until X. Turn left before Y on the corner''} These examples demonstrate how distinct previous steps are interrelated. 

To conclude, consistent with recent literature highlighting that providing a structured plan or decomposing tasks allows \acp{LLM} to better comprehend the environment and task structure~\cite{Ma_2024_CVPR, wei-etal-2025-plangenllms}, our findings confirm that \ac{LLM} reasoning is significantly enhanced when underpinned by effective planning mechanisms.

\subsection{POI detection}\label{sec:poi_detection}

\begin{table}[t!]
    \caption{Entity recognition performance across models evaluated on 100 instructions with an average of 3.82 annotated entities per instruction (382 total).}
    \centering
    \renewcommand{\arraystretch}{1}
    \resizebox{\linewidth}{!}{
    \begin{tabular}{>{\centering\arraybackslash}l|cc}
        \toprule
        \textbf{Model}           & \textbf{Num. Errors $\downarrow$} & \textbf{Correctness Rate $\uparrow$} \\
        \hline
        BERT     & 191 & 50.00 \%          \\
        GLiNER   & 137 & 64.13 \%          \\
        Gemini-3 Pro       & \textbf{4}   & \textbf{98.95 \%} \\
        \bottomrule
        \end{tabular}}
\label{tab:poi_ner}
\end{table}

As detailed in the Annotation section, we annotated the navigation instructions and obtained a total of 382 entities tagged as \acp{POI}. To assess the capability of different \ac{NER} models in this specific domain, we conducted a comparative analysis using three distinct approaches. We evaluated GLiNER~\cite{zaratiana2023gliner}, a generalist model designed for zero-shot named entity recognition, and a BERT-large~\cite{tjong-kim-sang-de-meulder-2003-introduction,DBLP:journals/corr/abs-1810-04805} model fine-tuned for standard NER tasks. Additionally, we tested the zero-shot annotation capabilities of Gemini-3 Pro.

Regarding the dataset characteristics, the average length of the selected navigation instructions is 49.46 words. The density of entities varies per instruction, with a minimum of one and a maximum of eight \acp{POI} per instruction. The most frequent \acp{POI} are Light, Starbucks, Duane Reade, Chase Bank, Subway, and Garden.
Table~\ref{tab:poi_ner} summarizes the quantitative results of this evaluation. 

Remarkably, Gemini failed to detect only 4 entities out of the total 382, which corresponds to an error rate of just 1.05\%. This performance is highly promising for this task because it indicates the model can operate at the level of a human expert. For the GLiNER setup, we utilized the ``urchade/gliner\_large-v2.1''\footnote{\url{https://huggingface.co/urchade/gliner_large-v2.1}} checkpoint with the following set of target labels: ``amenity'', ``cuisine'', ``leisure'', ``tourism'', ``shop'', ``highway'', and ``transportation''. Although the model is computationally efficient and the inference is very fast, its zero-shot performance remains significantly lower than the results achieved by Gemini. However, the annotation outputs suggest that GLiNER could potentially reach a near-perfect level with a small amount of fine-tuning.

In contrast, the results from the BERT model (``dslim/bert-large-NER''\footnote{\url{https://huggingface.co/dslim/bert-large-NER}}) were unsatisfactory. The model could detect standard store names but failed to identify a large portion of the domain-specific entities. Based on the error analysis, BERT misses approximately 2 \acp{POI} per instruction. Given that the average density of entities is 3.82 per instruction, this error rate is too high to be considered reliable for this application.

\subsection{Thinking Level}\label{sec:thinking_level}
\begin{table*}[t]
\caption{Performance comparison on Navigation Execution for different Thinking Level configurations categorized by instruction difficulty.}
\label{tab:thinking_performance}
\centering
\resizebox{\linewidth}{!}{%
\begin{tabular}{l||cccc|cccc|cccc}
\toprule
\multirow{3}{*}{\shortstack{\textbf{Thinking} \\ \textbf{Level}}} 
& \multicolumn{12}{c}{\textbf{Navigation Execution}} \\ 
\cmidrule(lr){2-13} 
& \multicolumn{4}{c|}{\textbf{Easy}} 
& \multicolumn{4}{c|}{\textbf{Medium}} 
& \multicolumn{4}{c}{\textbf{Hard}} \\
\cmidrule(lr){2-5} \cmidrule(lr){6-9} \cmidrule(lr){10-13} 
& \textbf{NE $\downarrow$} & \textbf{SR$\uparrow$} & \textbf{OSR$\uparrow$} & \textbf{nDTW$\uparrow$} 
& \textbf{NE $\downarrow$} & \textbf{SR$\uparrow$} & \textbf{OSR$\uparrow$} & \textbf{nDTW$\uparrow$} 
& \textbf{NE $\downarrow$} & \textbf{SR$\uparrow$} & \textbf{OSR$\uparrow$} & \textbf{nDTW$\uparrow$} \\
\midrule
Low & 46.6 & 73.5\% & 81.6\% & 0.721 & \colorbox{purple_color!7}{\textbf{42.1}} & 68.4\% & \colorbox{purple_color!7}{\textbf{81.6\%}} & \colorbox{purple_color!7}{\textbf{0.638}} & \colorbox{purple_color!7}{\textbf{86.1}} & \colorbox{purple_color!7}{\textbf{53.8\%}} & \colorbox{purple_color!7}{\textbf{61.5\%}} & 0.525 \\
High & \colorbox{purple_color!25}{\textbf{28.2}} & \colorbox{purple_color!25}{\textbf{81.6\%}} & \colorbox{purple_color!25}{\textbf{89.8\%}} & \colorbox{purple_color!25}{\textbf{0.801}} & 52.5 & \colorbox{purple_color!7}{\textbf{71.1\%}} & \colorbox{purple_color!7}{\textbf{81.6\%}} & 0.633 & \colorbox{purple_color!25}{\textbf{78.1}} & \colorbox{purple_color!25}{\textbf{61.5\%}} & \colorbox{purple_color!25}{\textbf{76.9\%}} & \colorbox{purple_color!25}{\textbf{0.602}} \\
\midrule
Auto & \colorbox{purple_color!7}{\textbf{35.6}} & \colorbox{purple_color!7}{\textbf{77.6\%}} & \colorbox{purple_color!7}{\textbf{85.7\%}} & \colorbox{purple_color!7}{\textbf{0.762}} & \colorbox{purple_color!25}{\textbf{30.9}} & \colorbox{purple_color!25}{\textbf{76.3\%}} & \colorbox{purple_color!25}{\textbf{84.2\%}} & \colorbox{purple_color!25}{\textbf{0.715}} & 93.3 & \colorbox{purple_color!7}{\textbf{53.8\%}} & \colorbox{purple_color!7}{\textbf{61.5\%}} & \colorbox{purple_color!7}{\textbf{0.527}} \\
\bottomrule
\end{tabular}
}
\small\textit{Note:} \colorbox{purple_color!25}{\textbf{Dark purple}} and \colorbox{purple_color!7}{\textbf{light purple}} indicate the top-performing and second top-performer per column. \\
\vspace{-8pt}
\end{table*}

\begin{table*}[t]
\caption{Overall performance and efficiency comparison across different thinking levels.}
\label{tab:overall_performance}
\centering
\resizebox{\linewidth}{!}{%
\begin{tabular}{l||cccc|cccc}
\toprule
\multirow{2}{*}{\textbf{Method}} 
& \multicolumn{4}{c|}{\textbf{Navigation Performance}} 
& \multicolumn{4}{c}{\textbf{Efficiency Statistics}} \\
\cmidrule(lr){2-5} \cmidrule(lr){6-9}
& \textbf{NE $\downarrow$} & \textbf{SR$\uparrow$} & \textbf{OSR$\uparrow$} & \textbf{nDTW$\uparrow$} 
& \textbf{Avg Steps} & \textbf{Avg Total Tokens $\downarrow$} & \textbf{Avg Thoughts Tokens$\downarrow$} & \textbf{Avg Prompt Tokens $\downarrow$} \\
\midrule
Low & 50.0 & 69.0\% & 79.0\% & 0.664 & 5.51 & \colorbox{purple_color!25}{\textbf{33,236}} & \colorbox{purple_color!25}{\textbf{13,010}} & \colorbox{purple_color!7}{\textbf{19,956}} \\
High & \colorbox{purple_color!7}{\textbf{43.9}} & \colorbox{purple_color!25}{\textbf{75.0\%}} & \colorbox{purple_color!25}{\textbf{85.0\%}} & \colorbox{purple_color!7}{\textbf{0.711}} & 5.50 & \colorbox{purple_color!7}{\textbf{41,347}} & \colorbox{purple_color!7}{\textbf{21,146}} & \colorbox{purple_color!25}{\textbf{19,933}} \\
\midrule
Auto & \colorbox{purple_color!25}{\textbf{41.3}} & \colorbox{purple_color!7}{\textbf{74.0\%}} & \colorbox{purple_color!7}{\textbf{82.0\%}} & \colorbox{purple_color!25}{\textbf{0.714}} & 5.76 & 43,700 & 22,335 & 21,098 \\
\bottomrule
\end{tabular}
}
\small\textit{Note:} \colorbox{purple_color!25}{\textbf{Dark purple}} and \colorbox{purple_color!7}{\textbf{light purple}} indicate the top-performing (or most efficient) and second-best method per column. \\
\vspace{-8pt}
\end{table*}

\begin{figure}
    \centering
    \includegraphics[width=\linewidth]{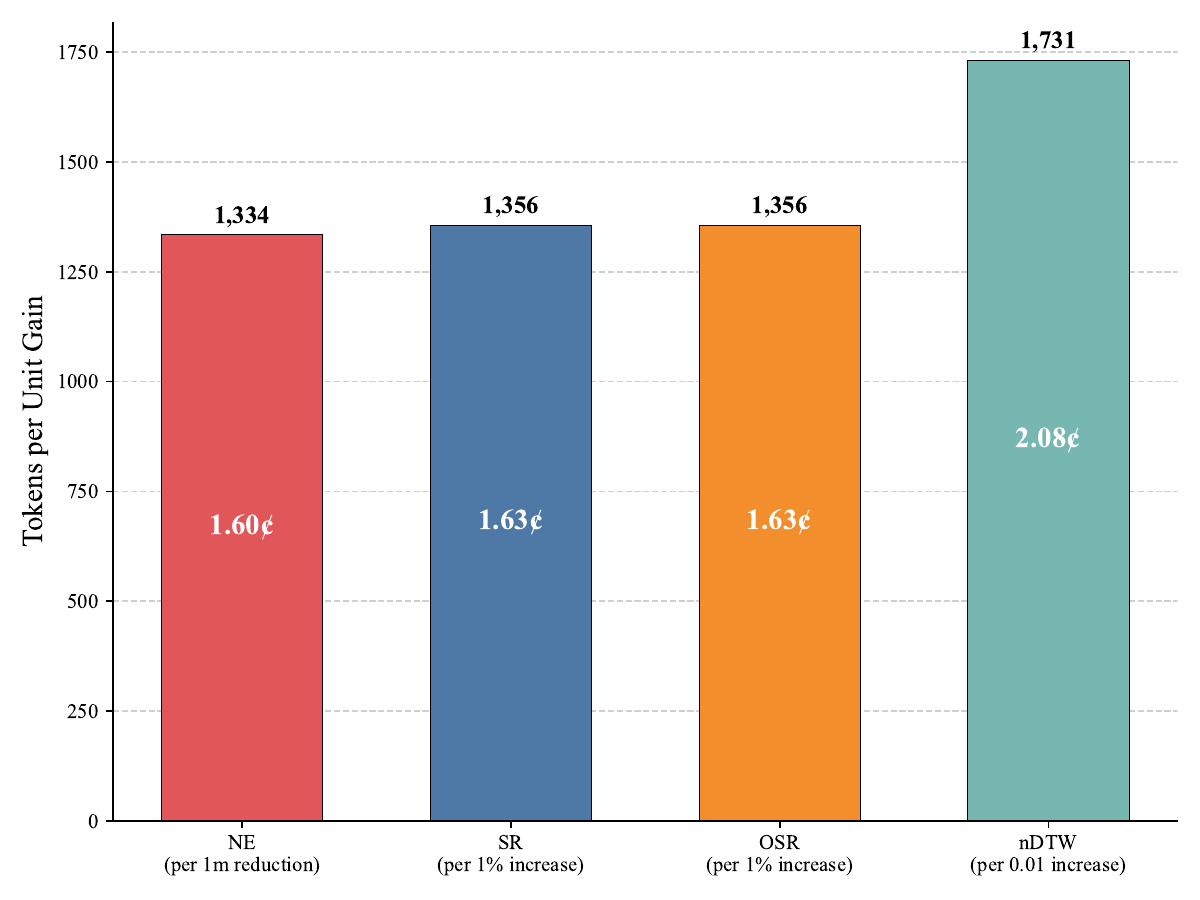}
    \caption{Marginal cost of performance improvements across navigation metrics. The additional computational cost required to achieve a single unit of improvement for different metrics. Costs are measured in thought tokens and estimated price in cents (center label, based on \$12 per 1M tokens) incurred over the baseline method.}
    \label{fig:marginal_cost}
\end{figure}

In the context of \acp{LLM}, when we talk about ``thinking'', we usually refer to simulated reasoning rather than consciousness or human-like cognition. It is a specific technical process where the model generates intermediate steps to solve a problem, rather than jumping directly to an answer. In a standard LLM, the model tries to predict the final answer immediately. 
This technique is called \ac{CoT}. So, we force the model to generate text that breaks down the problem, and the model effectively gives itself space to store intermediate variables and logic. This prevents the model from having to do all the computation in a single forward pass.

It is also worth noting the technical debate concerning whether this process constitutes \textit{``real'' thinking.} From a skeptic's perspective, the output is merely a statistically likely token prediction where the model possesses neither intent nor awareness. In contrast, from the functionalist view, if the model engages with a complex novel problem by breaking it down, identifying errors, and correcting them to produce an accurate solution, the process is functionally indistinguishable from reasoning regardless of the underlying biological substrate.

As documented in the Gemini API reference\footnote{\url{https://ai.google.dev/gemini-api/docs/thinking}}, the Gemini-3 series models utilize an internal "thinking process" that significantly improves their reasoning and multi-step planning abilities. This capability makes them highly effective for performing complex tasks. Consequently, we investigated how the thinking process influences the performance of our navigator agent. For this purpose, we evaluated the Gemini-3 Pro model on this task using three different configurations of the thinking level: low, high, and the default automatic setting without a defined limit.

Table \ref{tab:thinking_performance} details the Navigation Execution performance across varying instruction difficulties. We observe a non-linear relationship between the thinking level and task complexity. For \textbf{Easy} instructions, the \texttt{High} thinking configuration achieves the lowest Navigation Error (NE) of 28.2m and the highest Success Rate (SR) of 81.6\%. Similarly, in the \textbf{Hard} category, the \texttt{High} setting proves essential, attaining a dominant SR of 61.5\% compared to both \texttt{Low} and \texttt{Auto} configurations. This suggests that complex spatial reasoning tasks necessitate the extended intermediate computation provided by higher thinking budgets.

The aggregate performance, presented in Table \ref{tab:overall_performance}, highlights the distinct trade-off between efficacy and computational expenditure. While the \texttt{High} and \texttt{Auto} methods dominate the effectiveness metrics -- with \texttt{High} achieving the peak overall Success Rate of 75.0\% -- the \texttt{Low} setting functions as a highly resource-efficient baseline. It consumes approximately 33,236 total tokens on average, which is significantly lower than the 41,347 tokens utilized by \texttt{High} and 43,700 by \texttt{Auto}. The majority of this discrepancy arises from the ``Thoughts Tokens'' volume, which nearly doubles between the \texttt{Low} and \texttt{Auto} settings. Consequently, while increasing the thinking budget correlates with improved navigation outcomes, it incurs a substantial overhead in token consumption.

We analyzed the marginal cost of improvement as illustrated in Figure \ref{fig:marginal_cost}. This analysis reveals the additional computational investment required to achieve a single unit of gain across different metrics. We find that reducing the Navigation Error (NE) by 1 meter costs approximately 1,334 tokens (estimated at 1.60\textcent), whereas increasing the Success Rate (SR) by 1\% demands 1,356 tokens (1.63\textcent). The cost curve is steepest for trajectory fidelity; achieving a 0.01 increase in nDTW requires 1,731 tokens (2.08\textcent), significantly more than other metrics.

These findings suggest that the internal "thinking process" of the Gemini-3 model adheres to a law of diminishing returns. While the \texttt{High} thinking level is indispensable for maximizing success in complex scenarios (Hard instructions), the cost per unit of improvement is non-trivial. For applications where strict trajectory adherence (high nDTW) is less critical than simple goal arrival, the \texttt{Low} configuration may offer a more favorable cost-benefit ratio.

\section{Prompt Templates}\label{sec:prompt_templates}
We provide detailed prompt templates for each sub-task within \approach. These templates are designed to clearly define the objectives and input-output requirements for the models to ensure consistency and reproducibility across evaluations. Each template includes specific instructions and rigorous data schemas tailored to the corresponding sub-task, as illustrated in Prompts~\ref{prompt:extraction} and ~\ref{prompt:nav_agent}. For the Sub-instruction \& \ac{POI} Extraction task, the input is natural language text, which must be parsed into a structured JSON object containing sequential sub\_goals, status flags (e.g., ToDo), and categorized landmarks (\ac{OSM} \acp{POI}). On the other side, the Topological Graph Navigation task requires a navigation\_context input -- comprising current\_position, nodes, intersections, and connections -- from which the model must deduce a specific Target\_Node\_ID string and update the SubPlan\_Status.

\noindent 
\small
\refstepcounter{promptexample}\label{prompt:extraction}
\begin{tcolorbox}[ 
    enhanced, 
    colback=prompt1_bg, 
    colframe=prompt1_frame, 
    boxrule=2pt, 
    arc=4mm, 
    width=\linewidth, 
    title={\textbf{Prompt \thepromptexample: Sub-instruction \& \ac{POI} Extraction}}, 
    coltitle=white, 
    fonttitle=\bfseries, 
    attach boxed title to top left={yshift=-3mm, xshift=6mm}, 
    boxed title style={colback=prompt1_frame, arc=3mm}, 
    breakable,
    after skip=0.2cm,
] 
    \vspace{0.2cm} 
     
        \textbf{[System Role]}\;
        You are a Navigation Instruction Parser. Your goal is to translate natural language navigation instructions into structured, machine-readable sub-goals compatible with \ac{OSM} data.
    
        \vspace{0.2cm} 
        \textbf{[Definitions]}\\ 
        \textcolor{action_color}{1. LANDMARKS (OSM POIs):} Identify physical entities visible on a map.\\
        \texttt{- Traffic Control:} Traffic lights, stop signs. \\ 
        \texttt{- Amenities:} Banks, shops, restaurants, pharmacy, gas stations, bicycle rental, cinema. \\ 
        \texttt{- Natural:} Parks, etc. \\
        
        \textcolor{action_color}{2. ACTIONS:} Use only these verbs. \\
        \texttt{- MOVE\_FORWARD} (continue straight) \\ 
        \texttt{- TURN\_LEFT} \\ 
        \texttt{- TURN\_RIGHT}\\ 
        
        \textcolor{action_color}{3. RELATIONS:} Define the spatial relationship between the Action and the Landmark (e.g., \textit{"turn left AT the lights"}, \textit{"walk PAST the bank"}). \\
        
        \textbf{[Task]}\;
        Decompose the Full Instruction into a JSON object containing: \\
        1. A list of all unique ``landmarks'' mentioned. \\
        2. A sequential list of ``sub\_goals''. \\
        3. For each sub-goal, assign a ``status'' (TODO, IN\_PROGRESS, COMPLETED). Note: Unless live telemetry is provided, default all future steps to TODO. \\
        
        \textcolor{gray}{[OUTPUT FORMAT - STRICT JSON]} \\

        \textbf{Navigation Instruction:} Turn right at the light immediately in front $\ldots$
    
\end{tcolorbox}

\lstdefinelanguage{json}{
    basicstyle=\normalfont\ttfamily,
    stringstyle=\color{blue},
    numbers=left,
    numberstyle=\scriptsize,
    stepnumber=1,
    numbersep=8pt,
    showstringspaces=false,
    breaklines=true,
    frame=lines,
    backgroundcolor=\color{gray!10},
    literate=
     *{:}{{{\color{red}{:}}}}{1}
      {,}{{{\color{red}{,}}}}{1}
      {\{}{{{\color{black}{\{}}}}{1}
      {\}}{{{\color{black}{\}}}}}{1}
      {[}{{{\color{black}{[}}}}{1}
      {]}{{{\color{black}{]}}}}{1},
}

\noindent
\small
\refstepcounter{promptexample}\label{prompt:nav_agent}
\begin{tcolorbox}[ 
    enhanced, 
    colback=prompt1_bg, 
    colframe=prompt1_frame, 
    boxrule=2pt, 
    arc=4mm, 
    width=\linewidth, 
    title={\textbf{Prompt \thepromptexample: Topological Graph Navigation Agent}}, 
    coltitle=white, 
    fonttitle=\bfseries, 
    attach boxed title to top left={yshift=-3mm, xshift=6mm}, 
    boxed title style={colback=prompt1_frame, arc=3mm}, 
    breakable,
] 
    \vspace{0.2cm} 
     
    \textbf{[Task Description]}\;
    You are an embodied agent navigating using a topological graph-based map. Your goal is to determine the final target node for the current Sub-Goal based on the provided JSON navigation context.

    \vspace{0.2cm} 
    \textbf{[Input Format]}\\ 
    \texttt{Instruction:}  Go straight to the light and turn left. Proceed to the next $\ldots$, Trestle on Tenth should be on that near $\ldots$ \\
    \texttt{Current Sub-Goal:} Proceed to the next light and turn right. \\
    \texttt{Sub-Goal State:} \texttt{IN PROGRESS} \\
    \texttt{Landmarks} : light \\
    
    \textbf{Navigation Context (JSON):}
    \begin{lstlisting}[language=json]
{
  "navigation_context": {
    "current_position": { 
        "node_id": "cb04", "heading": 209, "compass_direction": "Southwest"
    },
    "previous_path": [{
        "node_id": "8b2f",
        "to_next": {
          "direction": "Forward", "heading": 298.5
        }
      },
      "..."
    ],
    "current_path_nodes": [{
        "node_id": "0cc7a",
        "to_next": {
          "direction": "Forward", "heading": 124.5
        }
      },
      "..."
    ]
  }
}
    \end{lstlisting}
    Includes `current\_position`, `nodes`, `connections`, `intersections`, and `pois`. \\

    \textcolor{gray}{[OUTPUT FORMAT - STRICT JSON]} \\
    
    \textbf{1. SubPlan\_Status}: \\
    \hspace*{2em}\texttt{- ``COMPLETED''}: If the Target\_Node\_ID you identified successfully finishes the specific action described in Current Sub-Goal. (Ignore future steps in the main Instruction). \\        
    \hspace*{2em}\texttt{- ``IN\_PROGRESS''}: If the Target\_Node\_ID is just an intermediate waypoint and you have not yet reached the location/intersection required by the Current Sub-Goal. \\
    \textbf{2. Next\_Place}: Target\_Node\_ID (String). The final Node ID AFTER executing the entire sub-goal instruction. \\

    \textbf{Planning State}:\\ 
    1. Go straight to the light and turn left. (\texttt{COMPLETED})\\
    2. Proceed to the next light and turn right. (\texttt{IN\_PROGRESS}, Iteration 1)\\
    3. Go all the way to the end of the block at the next light and stop in the middle of that intersection. (\texttt{TODO})\\

    \textbf{[Constraints]} \\
    - Use the provided JSON graph topology. Do not hallucinate coordinates or nodes not present in the ``nodes'' list. \\
    - Valid Movement: You can only move between nodes if they are explicitly linked in the ``connections'' list of the current node. \\
    - Node Types: \\
    \hspace*{2em}\texttt{- ``waypoint''}: A standard road segment. Usually has a ``Forward'' connection. \\
    \hspace*{2em}\texttt{- ``intersection''}: A decision point. Contains ``branches'' (Forward, Left, Right). \\
    - Evaluate SubPlan\_Status strictly against the Current Sub-Goal. \\
    - The output must be the final location for the current Sub-Goal. \\
    - POIs: Points of Interest are linked to specific nodes (`nearby\_node\_id`). Use the 'pois' list to locate landmarks.\\
    \hspace*{2em}\texttt{- ``On the Corner Validation''}: If an instruction specifies a turn at a landmark "on the corner," you must verify that the landmark's nearby\_node\_id is immediate to the intersection node (distance $3$ 15m or adjacent connection).\\
    \hspace*{2em}\texttt{- ``Turn Prevention''}: If the Landmark is visible in the pois list but its nearby\_node\_id requires moving Forward through the current intersection to reach it, DO NOT TURN. You must proceed IN\_PROGRESS towards the landmark. \\

    \textbf{[Clarifications]} \\
    - Processing Instructions: If the instruction is ``Go straight'', traverse through connected ``waypoint'' nodes until you reach an ``intersection'' or the max depth of the current graph, and if the instruction is \textit{``Turn [direction] at [landmark/intersection]''}: \\
    \hspace*{2em} - Identify the node associated with the landmark (from `pois`) or the next `intersection` node. \\
    \hspace*{2em} - From that intersection node, select the connection matching the direction (Left/Right). \\
    - Intersection Logic: When the instruction implies turning at an intersection, your target is the first node immediately after the turn. Look at the `intersection` node's `connections` or `branches`. Find the `target\_node\_id` corresponding to the requested `direction` (e.g., 'Right'). This `target\_node\_id` is your destination. \\
    - Landmark Logic: \\
    \hspace*{2em}\texttt{- ``Stopping Criteria''}: When a landmark is the destination, determine the target node based on the spatial preposition used (e.g., "past," "before," "at"). Select the first node that satisfies this relationship relative to the landmark's position.\\
    \hspace*{2em}\texttt{- ``Conditional Visibility''}: If an instruction requires a turn "regarding [landmark]" continue traversing nodes (implicitly "Go straight") until the landmark is confirmed visible. Do not execute the turn logic until this condition is met. \\
\end{tcolorbox}

\end{document}